\documentclass{article}

    \PassOptionsToPackage{numbers, compress}{natbib}

\usepackage[preprint]{neurips_2026}

\usepackage[utf8]{inputenc} 
\usepackage[T1]{fontenc}    
\usepackage{hyperref}       
\usepackage{url}            
\usepackage{booktabs}       
\usepackage{amsfonts}       
\usepackage{nicefrac}       
\usepackage{microtype}      
\usepackage[table]{xcolor}  
\usepackage{graphicx}       
\usepackage{amsmath}        
\usepackage{amssymb}        
\usepackage{listings}       
\usepackage{capt-of}        

\title{SceneCode: Executable World Programs for Editable Indoor Scenes with Articulated Objects}

\author{%
  \normalfont
  \parbox{0.94\textwidth}{\centering
    \textbf{Puyi Wang\textsuperscript{*1},
    Yuhao Wang\textsuperscript{*2,3},
    Linjie Li\textsuperscript{4},
    Zhengyuan Yang\textsuperscript{4}}\\[0.25em]
    \textbf{Kevin Qinghong Lin\textsuperscript{5},
    Yangguang Li\textsuperscript{1},
    Yu Cheng\textsuperscript{1}}\\[0.85em]
    \begin{tabular}{c}
      \textsuperscript{1}The Chinese University of Hong Kong\\
      \textsuperscript{2}Shanghai Jiao Tong University \quad
      \textsuperscript{3}Shanghai AI Laboratory\\
      \textsuperscript{4}Microsoft \quad
      \textsuperscript{5}University of Oxford\\
      \textsuperscript{*}Equal contribution.
    \end{tabular}
  }
}

\begin{document}

\maketitle

\begin{figure}[h]
  \centering
  \includegraphics[width=\linewidth]{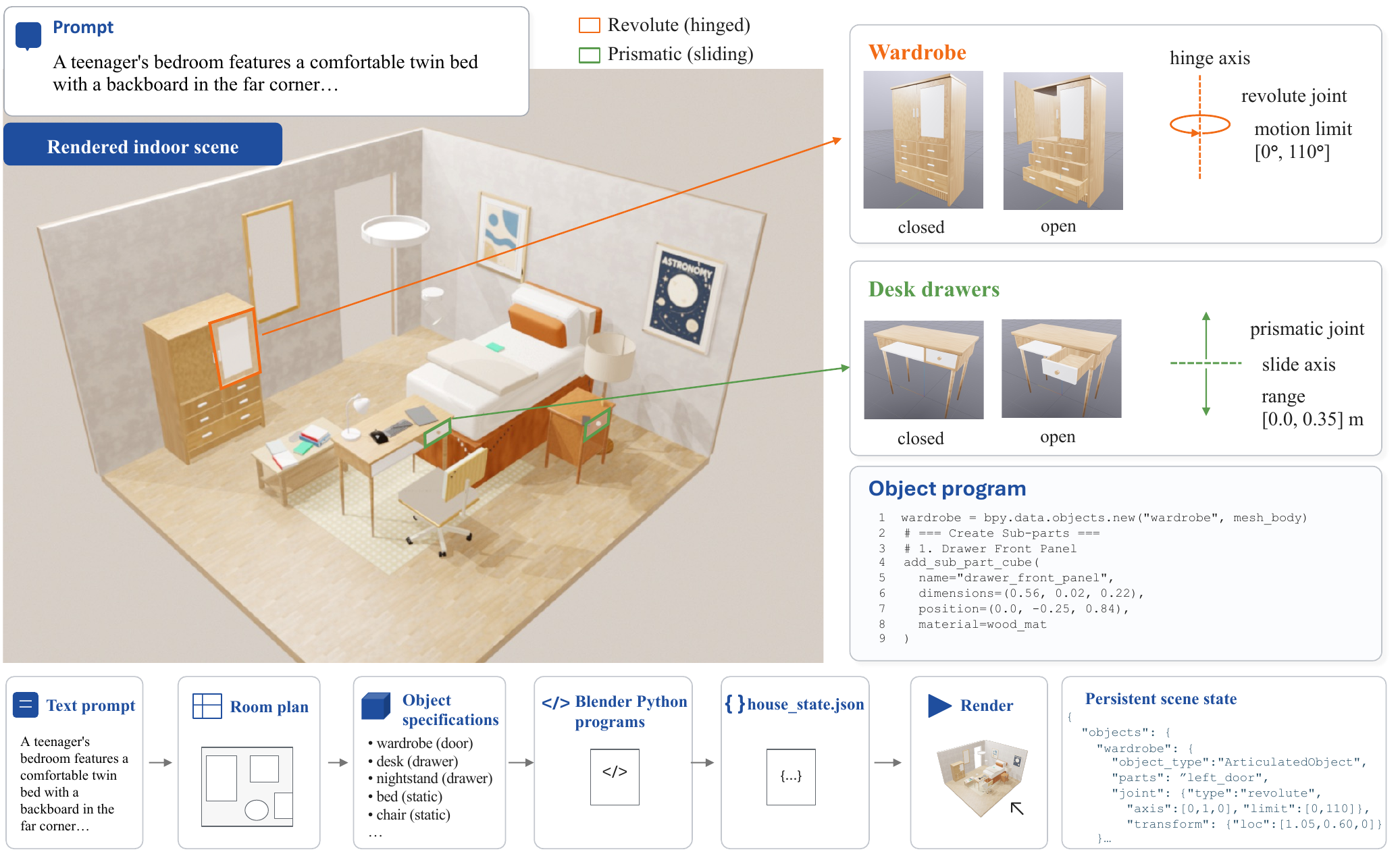}
  \caption{\textbf{Overview of \textsc{SceneCode}.} Given a natural language scene prompt, our framework compiles it into an executable, code-driven indoor scene with interactable objects.}
  \label{fig:teaser}
\end{figure}


\begin{abstract}
Indoor scene synthesis underpins embodied AI, robotic manipulation, and simulation-based policy evaluation, where a useful scene must specify not only what the environment looks like, but also how its objects are structured. Existing pipelines, however, typically represent generated content as static meshes and inherit articulation only from curated asset libraries, which limits object-level controllability and prevents new interactable assets from being produced on demand. We address this gap by formulating physically interactable indoor scene synthesis as programmatic world generation, and present \textbf{SceneCode}, a framework that compiles a natural language prompt into an executable, code-driven indoor world rather than a collection of opaque meshes. A room-level agentic backbone first turns the prompt into a structured house layout and emits per-object \emph{AssetRequest}s through a planner--designer--critic loop. Each request is then routed to one of five code-generation strategies and converted into a synthesized part-wise Blender Python programs that are validated through an execution-guided repair-and-refine loop. The resulting programs are compiled into simulation-ready assets, and exported as SDF for physics simulation. A persistent scene-state registry links object requests, executable programs, rendered geometry, and simulation assets, turning scene assembly into a traceable and locally editable world-building process. 
We evaluate SceneCode across scene-level synthesis, object-level asset quality, human judgment, and downstream robot interaction. 
Results show that executable world programs improve prompt-faithful indoor scene generation and produce assets with cleaner mesh structure, and simulator-loadable articulation metadata.
Project page: \url{https://scene-code.github.io/}.
\end{abstract}

\section{Introduction}
\label{sec:intro}

Indoor scene synthesis is a fundamental substrate for embodied AI~\citep{kolve2017ai2thor,savva2019habitat}, robotic manipulation~\citep{mu2021maniskill,nasiriany2024robocasa}, and simulation-based policy evaluation~\citep{szot2021habitat2,li2023behavior}.
By generating diverse indoor environments, such systems can provide scalable virtual worlds for training agents, testing manipulation skills, and collecting synthetic interaction data without expensive manual modeling~\citep{deitke2022procthor,raistrick2023infinigen}.
Therefore, the goal of indoor scene synthesis is not merely to create a visually plausible room composed of well-arranged objects.
For an embodied agent, an indoor scene must expose physical structure and interaction mechanisms.
Thus, a useful generated scene should specify not only what the environment looks like, but also how its objects are structured, how they move, and how agents can physically act upon them.

Existing methods have advanced this goal from different directions.
Retrieval-based and LLM-guided systems can populate diverse environments with large asset libraries~\citep{feng2023layoutgpt,yang2024holodeck}, layout-centric methods improve spatial plausibility through optimization~\citep{paschalidou2021atiss,lin2024instructscene,sun2025layoutvlm}, and recent agentic systems generate simulation-ready environments with dense object populations and physical properties~\citep{pfaff2026scenesmith,pun2025hsm}.
Nevertheless, most pipelines still represent generated content as static meshes.
Even when articulated objects are present, their part structure and joint semantics are typically inherited from curated datasets~\citep{xiang2020sapien,wang2022adaafford} rather than generated as part of the scene representation.
That is, the choice of interactable objects is constrained: such objects cannot be customized on demand and face the problem that, if a given object is absent from the dataset, it simply cannot be retrieved.
This limits object-level controllability and the scalable generation of new interactable assets.

To address this challenge, we formulate physically interactable indoor scene synthesis as programmatic world generation and propose \textbf{SceneCode}: a framework that generates indoor scenes as executable programs rather than static visual assets.
As illustrated in Figure~\ref{fig:teaser}, SceneCode exposes a generated scene at multiple levels: a renderable room, a persistent scene state, and object-level programs with explicit parts and interaction mechanisms.
Code provides a natural representation for interactable scenes because it can make object geometry, part decomposition, material assignment, physical attributes, and motion mechanisms explicit in a unified form.
This representation also aligns well with the emerging capability of vision-language models (VLMs)~\citep{openai2023gpt4v,liu2023llava,team2023gemini} to generate structured programs from natural language specifications~\citep{chen2021codex,liang2023code,singh2023progprompt}.
In this way, a 3D object is not only generated as a visually plausible piece of furniture, but as a structured object with controllable states.
By making interaction an intrinsic part of the generated program, SceneCode enables interactable objects to be generated on demand rather than selected only from curated articulated asset libraries or produced through laborious manual modeling, and provides a foundation for physically grounded indoor scene synthesis.

We instantiate SceneCode as an agentic text-to-scene pipeline that compiles a natural language prompt into an executable indoor world.
Specifically, given a prompt, the system first infers a room-level plan, including room geometry, semantic descriptions, object requirements, spatial constraints.
Instead of satisfying these object requirements by selecting assets from a fixed library or producing opaque meshes, SceneCode converts each requirement into a structured object specification and invokes a VLM-based program synthesizer to generate Blender Python code~\citep{raistrick2023infinigen,sun20233dgpt,hu2024scenecraft}.
The generated program builds the object part by part from geometric primitives, assigns materials and UVs to each semantic part, and attaches physical attributes, collision proxies, and prismatic or revolute joints where appropriate.
After execution, each object program is registered into a persistent \texttt{house\_state} file, which records layout, room geometry, object transforms, support surfaces, geometry paths, bounding boxes, and interaction metadata.
The final output is a scene with physically annotated, interactable objects that remain editable and locally regenerable, supporting constraint modification and downstream object-level interaction in simulation.


We evaluate SceneCode on 30 natural language prompts spanning six indoor scene categories, comparing against SceneSmith~\citep{pfaff2026scenesmith}, HSM~\citep{pun2025hsm}, and LayoutVLM~\citep{sun2025layoutvlm} at the scene level and SAM 3D Objects~\citep{meta2025sam3d} at the asset level. SceneCode achieves the best semantic fidelity among scene-level baselines, with the highest object-count and attribute scores, and also improves navigability, collision, and floor-containment metrics. Human raters judge SceneCode more prompt-faithful than each baseline within matched comparison groups. At the object level, SceneCode produces more usable assets than SAM 3D Objects~\citep{meta2025sam3d}. Finally, MuJoCo~\citep{todorov2012mujoco} demonstrations show that the generated articulated assets retain independent movable links and executable joints for contact-based robot interaction.

In summary, our key contributions are threefold:
\begin{itemize}
    \item We introduce \textbf{SceneCode}, an executable code representation for indoor scene synthesis, which explicitly captures layout and object attributes in code format.
    \item We propose a VLM-driven object synthesis procedure that generates household objects as explicit programs, enabling new interactable assets to be generated on demand rather than selected only from fixed articulated-object datasets.
    \item We evaluate SceneCode across scene-level synthesis, object-level asset quality, human judgment, and robot interaction, demonstrating prompt-faithful scene generation with interaction-ready articulated object assets.
\end{itemize}

\section{Related Work}
\label{sec:related}

\paragraph{Indoor Scene Synthesis.}
Learning-based scene synthesizers model object layout distributions from annotated room datasets via autoregressive transformers or denoising diffusion~\citep{paschalidou2021atiss,tang2024diffuscene}.
LLM- and retrieval-guided pipelines instead populate rooms by querying curated 3D libraries: representative works include Holodeck~\citep{yang2024holodeck}, LayoutVLM~\citep{sun2025layoutvlm}, HSM~\citep{pun2025hsm}, and the agentic SceneSmith~\citep{pfaff2026scenesmith}, which mixes dataset retrieval with image-to-3D generation for simulation-ready scenes; additional LLM/retrieval-based and procedural systems are surveyed in Appendix~\ref{app:more_related_work}.
In contrast, SceneCode synthesizes the objects themselves as executable programs, removing the dependency on a fixed asset library and exposing each object's parts and joints to the scene representation.

\paragraph{Code-Driven and Procedural 3D Generation.}
Programs offer a compact, editable representation of 3D content.
Infinigen~\citep{raistrick2023infinigen} uses hand-written procedural rules for photorealistic worlds, ShapeAssembly~\citep{jones2020shapeassembly} learns part-program priors over shapes, and recent VLM-driven systems such as SceneCraft~\citep{hu2024scenecraft} and MeshCoder~\citep{wang2025meshcoder} synthesize Blender Python from natural language or point clouds; further code-generation systems are discussed in Appendix~\ref{app:more_related_work}.
These efforts mostly target either single-object modeling or scene-level visual layout, with limited support for downstream physical interaction.
SceneCode extends program-based generation to interactable indoor scenes through a routed, verified \emph{ObjectPlan} that drives part-wise Blender programs and compiles into URDF/SDF assets registered into a persistent scene state.

\section{Method}
\label{sec:method}

Given a natural language scene prompt, SceneCode produces a renderable scene together with scene-state metadata and simulation-ready asset files.
Our system separates the problem of indoor scene synthesis into two coupled levels: a room-level agent determines what objects are needed and where they should be placed, while a code-driven asset generator determines how each object is constructed, and compiled into renderable and simulation-ready artifacts.

\begin{figure}[t]
  \centering
  \includegraphics[width=\linewidth]{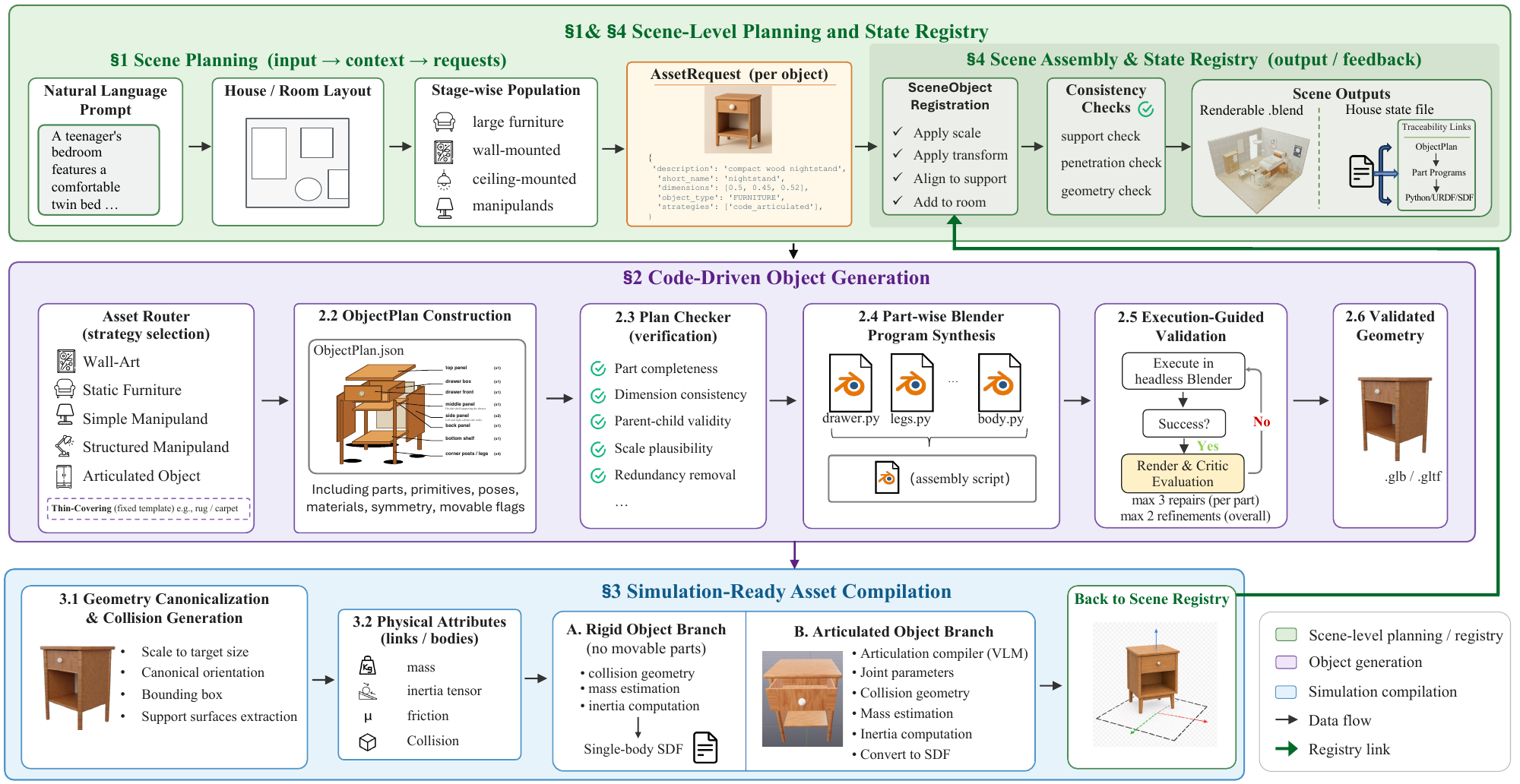}
  \caption{\textbf{Overview of \textsc{SceneCode}.} From room-level planning to code-driven object generation, simulation-ready compilation, and scene-state registration.}
  \label{fig:method}
\end{figure}

An overview of the full pipeline is illustrated in Figure~\ref{fig:method}.
We briefly introduce the room-level backbone that provides contextual object requests in Section~\ref{sec:room_backbone}, and focus on the construction of executable object programs in Section~\ref{sec:object_program}.
Next, we introduce the simulation-ready asset compilation in Section~\ref{sec:sim_compile}, and finally the scene assembly and state serialization in Section~\ref{sec:scene_assembly}.

\subsection{Room-Level Agentic Scene Backbone}
\label{sec:room_backbone}

The room-level backbone transforms a scene prompt into a set of per-room object specifications that drive subsequent program synthesis.
Concretely, it produces a structured house layout $\mathcal{H}$ together with an ordered sequence of object requests $\{\mathcal{A}^{(t)}\}_{t=1}^{T}$.
Within each room, requests are emitted in four semantic stages: large furniture, wall-mounted objects, ceiling-mounted objects, and manipulable items.
Each stage is driven by a planner--designer--critic loop $(\mathsf{Plan},\mathsf{Design},\mathsf{Critic})$: $\mathsf{Plan}$ selects the next placement task, $\mathsf{Design}$ invokes tools to create or modify objects, and $\mathsf{Critic}$ evaluates the intermediate scene from rendered views, scene-state information, and geometric consistency checks.
The output of stage $t$ is not a final asset but an \emph{AssetRequest}
\[
\mathcal{A} \;=\; (c,\;\tau,\;\mathbf{d},\;\sigma,\;\mathbf{T},\;\rho),
\]
specifying the object category $c$, textual description $\tau$, target dimensions $\mathbf{d}\in\mathbb{R}^{3}$, style context $\sigma$, placement transform $\mathbf{T}\in\mathrm{SE}(3)$, and support relation $\rho$.
The sequence $\{\mathcal{A}^{(t)}\}_{t=1}^{T}$ is the contract carried into the object-level program synthesis stage.

\subsection{Code-Driven Object Generation}
\label{sec:object_program}

This subsection turns each AssetRequest into an executable Blender program whose output is a part-decomposed, renderable mesh.
The pipeline proceeds through five steps: routing the request to a construction strategy, lifting it into a structured ObjectPlan, verifying the plan, synthesizing per-part Blender programs, and validating the resulting code through execution.

\paragraph{Asset Request and Strategy Routing.}
Directly prompting a single VLM to emit a Blender script from $\mathcal{A}$ is unreliable across diverse indoor objects, since different object families require different construction priors: wall art needs a thin canvas with an image material, whereas articulated objects must preserve movable components for downstream joint compilation.
SceneCode therefore introduces a router $r:\mathcal{A}\!\mapsto\!s$ that dispatches each request to one of five VLM-based code-generation strategies (\textsc{WallArt}, \textsc{StaticFurn}, \textsc{SimpleManip}, \textsc{StructManip}, \textsc{Artic}), or to a fixed code template \textsc{ThinCover} reserved for thin coverings (rugs, carpets) that bypasses free-form VLM synthesis.
The five VLM-based strategies cover the dominant construction priors of indoor objects:
\begin{itemize}
  \item \textsc{WallArt}: posters, framed artwork, and other print-like wall-mounted objects.
  \item \textsc{StaticFurn}: large rigid furniture without functional moving parts, such as beds, shelves, and sofas.
  \item \textsc{SimpleManip}: structurally simple rigid objects with a dominant shape, such as bowls and plates.
  \item \textsc{StructManip}: rigid objects with multiple visible components but no articulation, such as mugs and phones.
  \item \textsc{Artic}: objects with functional movable parts, such as cabinets and refrigerators, to be compiled into a link--joint structure.
\end{itemize}
Each VLM-based route is paired with a specialized construction prompt that encodes geometry-aware coding constraints for symmetry, repeated structures, and curve construction; for example, curved shapes are constructed from explicit sampled points and analytic primitives rather than unconstrained B\'ezier curves.
Full prompt listings are provided in Appendix~\ref{app:prompts}.

\paragraph{Reference-Conditioned ObjectPlan Construction.}
For every strategy except \textsc{ThinCover}, $\mathcal{A}$ is first lifted to a structured \emph{ObjectPlan}~$\mathcal{P}$ to reduce ambiguity in code synthesis.
A reference image $I_{\mathrm{ref}}$ is generated from the description--style pair $(\tau,\sigma)$, and an object planner consumes $(\mathcal{A},I_{\mathrm{ref}})$ to produce
\[
\mathcal{P} \;=\; \{(p_i,\;s_i,\;\mathbf{T}_i^{\mathrm{loc}},\;m_i,\;\pi_i,\;\mu_i)\}_{i=1}^{N},
\]
where $p_i$ is a semantic part, $s_i\in\{\textsc{box},\textsc{cyl},\textsc{sph},\textsc{torus},\textsc{curve}\}$ its primitive type, $\mathbf{T}_i^{\mathrm{loc}}$ its pose in the object-local frame, $m_i$ its material, $\pi_i$ its symmetry tag, and $\mu_i\in\{0,1\}$ a movability flag.
For requests routed to \textsc{Artic}, parts with $\mu_i=1$ (e.g., doors, drawers) are later compiled into a joint schema.


\paragraph{ObjectPlan Verification.}
Free-form plans may omit functional parts, propose implausible part scales, or place parts inconsistently with the object body, so we apply a checker $\mathcal{V}:\mathcal{P}\to\mathcal{P}^{\star}$ before code synthesis.
In practice, $\mathcal{V}$ combines lightweight rule-based validation with an LLM-based revision step, and targets four desiderata:
\begin{itemize}
  \item \emph{Part completeness}: redundant components are removed and missing functional parts are inserted with respect to the requested category $c$.
  \item \emph{Dimension plausibility}: implausible per-part scales $s_i$ are corrected so that the parts remain consistent with the requested category and the target dimensions $\mathbf{d}$.
  \item \emph{Spatial consistency}: the local poses $\{\mathbf{T}_i^{\mathrm{loc}}\}$ are revised so that parts respect the object-local frame and integrate coherently with the body.
  \item \emph{Movable-part independence}: parts with $\mu_i=1$ are kept as separately addressable components rather than fused with their parent, which is the precondition for the downstream joint compilation in Section~\ref{sec:sim_compile}.
\end{itemize}
The verified plan $\mathcal{P}^{\star}$ then serves as the contract that subsequent code generation must satisfy.

\paragraph{Part-wise Blender Program Synthesis.}
Given $(\mathcal{P}^{\star},I_{\mathrm{ref}})$, a part constructor synthesizes one Blender Python program $\pi_i$ per part, returning a primitive-based mesh $\mathcal{M}_i$ in the object-local frame together with procedural materials.
A composition script then assembles the object mesh $\mathcal{M}$ by unioning the part meshes, keeping each $\mathcal{M}_i$ as a separately named Blender object so that movable and non-movable components remain semantically decomposable rather than being fused into a single opaque mesh.
Complete part-level code listings are provided in Appendix~\ref{app:code}.

\paragraph{Execution-Guided Program Validation.}
Each $\pi_i$ is executed in headless Blender and validated by a two-budget loop with $K_{\mathrm{exec}}=3$ and $K_{\mathrm{ref}}=2$:
\begin{enumerate}
  \item \emph{Execute}: run $\pi_i$ to materialize $\mathcal{M}_i$.
  \item \emph{Repair}: if execution fails, return the traceback together with the offending code to the synthesizer; up to $K_{\mathrm{exec}}$ repair attempts are allowed per part.
  \item \emph{Refine}: upon successful execution, a critic agent inspects rendered images of the assembled object and judges whether the requested category, structure, and material requirements are met, triggering up to $K_{\mathrm{ref}}$ refinement iterations.
\end{enumerate}
This execution-guided loop improves code reliability and prevents invalid assets from entering the scene-level assembly stage.

\subsection{Simulation-Ready Asset Compilation}
\label{sec:sim_compile}

This stage converts a generated visual object into simulator-compatible asset files, packaging rigid objects as single-body SDFs and articulated objects as link--joint structures with inferred joints.
Formally, SceneCode applies a compilation map $C$ that bridges visual geometry and physical interaction.
For rigid requests, $C:\mathcal{M}\mapsto\mathrm{SDF}$ produces a single-body asset with collision and inertial properties.
For articulated requests, $C:(\mathcal{P}^{\star},\mathcal{M})\mapsto(\mathcal{J},\mathrm{SDF})$ additionally returns a joint schema $\mathcal{J}$ inferred by a VLM-assisted articulation compiler over the parts of $\mathcal{P}^{\star}$ with $\mu_i=1$: for each movable part, the compiler emits a parent link, a joint type ($\textsc{revolute}$ for hinged or $\textsc{prismatic}$ for sliding), and a plausible joint origin, axis, and motion range, covering the two dominant indoor mechanisms.

To support contact-based interaction, each link $i$ is endowed with approximate physical attributes: a mass $m^{\mathrm{phy}}_i$ estimated from object- and part-level semantics, an inertia tensor $\mathbf{I}_i$ computed from $(\mathcal{M}_i,m^{\mathrm{phy}}_i)$, and a collision proxy $\mathcal{C}_i$ derived from their simplified geometric envelopes.
The resulting assets is exported as an SDF artifact for downstream physics simulation.

\subsection{Scene Assembly and State Serialization}
\label{sec:scene_assembly}

Scene assembly closes the loop between room-level planning and code-driven object generation, ensuring that the visual mesh, the executable program, and the simulation artifacts of every object remain linked through a shared identifier.
Concretely, each generated object is registered as a \emph{SceneObject} into a scene-level registry under a shared identifier $\mathrm{id}$ that links its content (e.g., request, programs, and joint schema if articulated).
Placement amounts to scaling $\mathcal{M}$ to the target dimensions $\mathbf{d}$, applying the planned transform $\mathbf{T}\in\mathrm{SE}(3)$, and aligning the object with its support relation $\rho$.
The shared $\mathrm{id}$ is what makes scene assembly traceable and locally editable: the rendered mesh, the executable programs $\{\pi_i\}$, and the simulation artifacts $(\mathcal{J},\mathrm{SDF})$ all reference the same object instance, enabling parameter-level edits and partial re-execution.

\section{Experiments}
\label{sec:experiments}

\subsection{Experimental Setup}
\label{sec:exp_setup}

\paragraph{Baselines.}
We compare {SceneCode} against three recent text-to-scene baselines: SceneSmith~\citep{pfaff2026scenesmith}, {HSM}~\citep{pun2025hsm}, and LayoutVLM~\citep{sun2025layoutvlm}.
SceneSmith is an agentic simulation-ready scene generation system, while {HSM} and LayoutVLM represent recent layout- and motif-oriented indoor scene generation approaches.
Together, these baselines cover complementary scene synthesis strategies, including agentic scene construction, hierarchical motif placement, and vision-language layout optimization.

\paragraph{Input text descriptions.}
We evaluate all methods on 30 room-level prompts selected from SceneEval-100~\citep{tam2026sceneeval}.
The prompt set spans six indoor room categories: bedroom, living room, dining room, kitchen, basement, and bathroom, ranging from short object-list descriptions to more detailed instructions.
The complete prompt list is provided in Appendix~\ref{app:evaluation_prompts}.

\paragraph{Automatic evaluation.}
We adopt the scene-level metrics from SceneEval~\citep{tam2026sceneeval}:
CNT (object count),
ATR (object attribute),
OOR (object--object relationship),
OAR (object--architecture relationship),
SUP (support),
ACC (accessibility),
NAV (navigability),
COL (collision),
OOB (out of bounds),
and OPC (opening clearance).
For object-level evaluation, we use a set of mesh- and material-level metrics that reflect downstream usability:
material slot count (MAT),
PBR channel coverage (PBR),
non-manifold edge count (NME),
total face count (FAC),
total vertex count (VTX),
and UV island count (UVI).
Detailed metric definitions are provided in Appendix~\ref{app:metric_details}.

\paragraph{User study.}
We conduct a user study with nine participants split evenly into three groups: Group A compares {SceneCode} against SceneSmith, Group B against {HSM}, and Group C against LayoutVLM.
We primarily assess prompt faithfulness, which directly reflects whether the generated scene follows the input description; additional preference and realism ratings are reported in Appendix~\ref{app:user_study}.
Because {SceneCode} is rated in all three groups while each baseline is rated only in its own group, we report the within-group difference $\Delta = \overline{\text{SceneCode}} - \overline{\text{Baseline}}$ to keep the comparison fair across raters and prompt subsets.

\subsection{Room-Level Scene Synthesis}
\label{sec:exp_scene}

\begin{figure}[t]
  \centering
  \includegraphics[width=\linewidth]{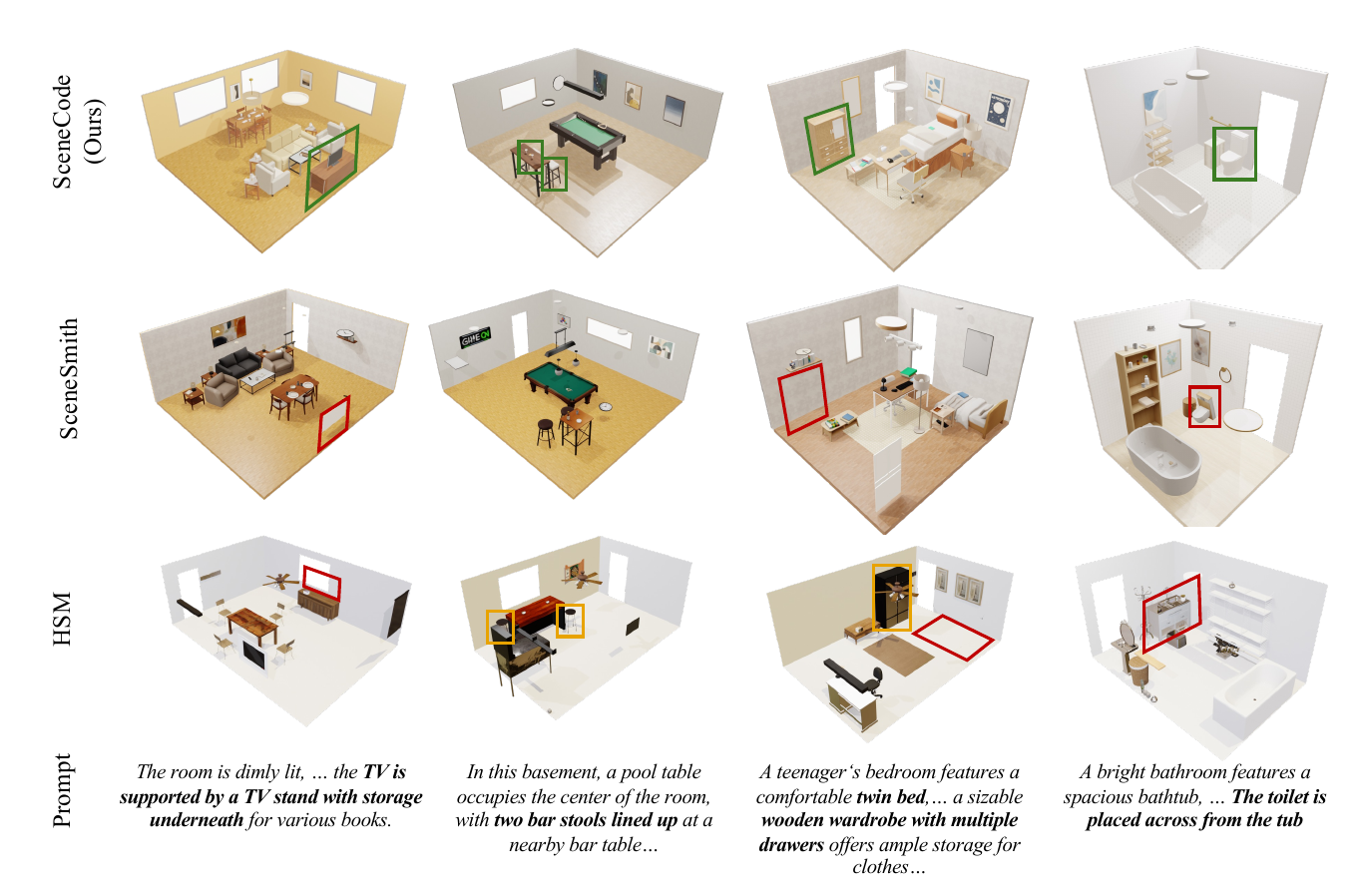}
  \caption{\textbf{Room-level qualitative comparison.} SceneCode shows better prompt fidelity than the baselines. See Appendix~\ref{app:more_scene_demos} for additional examples.}
  \label{fig:scene_results}
\end{figure}

\paragraph{Qualitative comparison.}
Figure~\ref{fig:scene_results} shows that {SceneCode} renders scenes that closely match the prompt's described atmosphere, object set, and spatial layout.
SceneSmith generates plausible furniture but cannot customize articulated objects.
HSM produces locally coherent placements within each motif subtree, but since its motifs operate inside individual subtrees, cross-subtree relations such as ``the desk faces the bed'' are not reliably realized.
The qualitative gap is consistent with the CNT and ATR advantages reported in Table~\ref{tab:room_level_results}.

\begin{table}[t]
  \centering
  \caption{\textbf{Room-level quantitative results.} We report scene-level SceneEval~\citep{tam2026sceneeval} metrics (in \%, with 95\% CI in smaller font next to the mean) on 30 room-level prompts. The last column ($\Delta$PF) reports the per-trial paired prompt-faithfulness difference SceneCode$-$baseline as a percentage of the 5-point rating scale. $\uparrow=$ higher is better, $\downarrow=$ lower is better. Best results per column are in \textbf{bold}.}
  \label{tab:room_level_results}
  \resizebox{\linewidth}{!}{
  \begin{tabular}{lccccccccccc}
    \toprule
    Method & CNT $\uparrow$ & ATR $\uparrow$ & OOR $\uparrow$ & OAR $\uparrow$ & SUP $\uparrow$ & ACC $\uparrow$ & NAV $\uparrow$ & COL $\downarrow$ & OOB $\downarrow$ & OPC $\downarrow$ & $\Delta$PF $\uparrow$ \\
    \midrule
    {HSM} & 53.3\,{\scriptsize $\pm$9.3} & 58.1\,{\scriptsize $\pm$14.7} & 22.3\,{\scriptsize $\pm$12.3} & 50.0\,{\scriptsize $\pm$10.4} & \textbf{70.9}\,{\scriptsize $\pm$7.2} & 85.8\,{\scriptsize $\pm$4.8} & 99.1\,{\scriptsize $\pm$1.0} & 18.4\,{\scriptsize $\pm$6.8} & 7.4\,{\scriptsize $\pm$3.6} & 10.9\,{\scriptsize $\pm$8.8} & {\color{green!60!black}+13.2\,{\scriptsize $\pm$4.4}} \\
    LayoutVLM & 64.2\,{\scriptsize $\pm$8.0} & 31.1\,{\scriptsize $\pm$13.7} & 27.2\,{\scriptsize $\pm$12.1} & 53.1\,{\scriptsize $\pm$14.8} & 25.7\,{\scriptsize $\pm$7.7} & \textbf{93.3}\,{\scriptsize $\pm$5.2} & 100.0\,{\scriptsize $\pm$0.0} & 21.1\,{\scriptsize $\pm$10.0} & 5.9\,{\scriptsize $\pm$3.5} & N/A & {\color{green!60!black}+24.6\,{\scriptsize $\pm$5.8}} \\
    SceneSmith & 78.8\,{\scriptsize $\pm$6.7} & 71.6\,{\scriptsize $\pm$9.4} & \textbf{37.2}\,{\scriptsize $\pm$13.3} & \textbf{72.4}\,{\scriptsize $\pm$10.3} & 66.7\,{\scriptsize $\pm$4.0} & 52.2\,{\scriptsize $\pm$8.7} & 97.8\,{\scriptsize $\pm$2.1} & 18.1\,{\scriptsize $\pm$4.8} & 0.6\,{\scriptsize $\pm$0.7} & \textbf{7.9}\,{\scriptsize $\pm$4.8} & {\color{green!60!black}+2.8\,{\scriptsize $\pm$4.0}} \\
    \rowcolor{black!8} {SceneCode} & \textbf{79.4}\,{\scriptsize $\pm$6.0} & \textbf{74.0}\,{\scriptsize $\pm$11.7} & 32.6\,{\scriptsize $\pm$12.6} & 61.1\,{\scriptsize $\pm$12.6} & 44.3\,{\scriptsize $\pm$12.8} & 74.2\,{\scriptsize $\pm$9.1} & \textbf{100.0}\,{\scriptsize $\pm$0.0} & \textbf{11.3}\,{\scriptsize $\pm$4.3} & \textbf{0.4}\,{\scriptsize $\pm$0.9} & 12.4\,{\scriptsize $\pm$6.7} & --- \\
    \bottomrule
  \end{tabular}
  }
\end{table}

\paragraph{Quantitative summary.}
{SceneCode} is the only method that simultaneously leads on semantic fidelity (CNT, ATR) and physical usability (NAV, COL, OOB): its per-object code realizes prompt attributes directly at construction time rather than approximating them via retrieved meshes, and the resulting clean, bounding-box-faithful geometry lets the placer reason precisely about collisions and floor containment.
The user study (last column) provides supportive evidence from human raters: {SceneCode} is judged more prompt-faithful than every baseline, with $\Delta\text{PF}=+2.8\%$, $+13.2\%$, and $+24.6\%$ over SceneSmith, HSM, and LayoutVLM respectively (on a 5-point scale), and the gap widens on the more layout-oriented baselines that lack an explicit attribute realization stage.
Per-metric numbers and analyses of the metrics on which {SceneCode} does not lead (OOR, OAR, SUP, ACC, OPC), as well as preference loss rates and realism ratings, are deferred to Appendix~\ref{app:more_analysis}.
Overall, programming objects rather than retrieving or shaping them produces scenes that are at once prompt-faithful, physically usable, and structurally coherent.

\subsection{Object-Level Geometry and UV Quality}
\label{sec:exp_object}

We compare {SceneCode} against the image-to-3D baseline SAM 3D Objects~\citep{meta2025sam3d}.
For a fair comparison, both methods are run on the same set of object requests issued by the room-level agent on the SceneEval-100 prompts, using the same reference images.
We evaluate each generated asset with the six mesh- and material-level metrics introduced in Section~\ref{sec:exp_setup} (MAT, PBR, NME, FAC, VTX, UVI); detailed definitions are provided in Appendix~\ref{app:object_level_metrics}.

\begin{table}[t]
  \centering
  \caption{\textbf{Object-level quantitative results.} Mesh- and material-level metrics averaged over the same set of object requests issued by the room-level agent on the SceneEval-100 prompts. Higher is better for $\uparrow$ metrics, lower is better for $\downarrow$ metrics. Best results are shown in \textbf{bold}.}
  \label{tab:object_level_results}
  \resizebox{0.8\linewidth}{!}{
  \begin{tabular}{lcccccc}
    \toprule
    Method & MAT $\uparrow$ & PBR $\uparrow$ & NME $\downarrow$ & FAC $\downarrow$ & VTX $\downarrow$ & UVI $\downarrow$ \\
    \midrule
    SAM 3D Objects~\citep{meta2025sam3d} & 1.0000 & 0.2000 & 0.0164 & 12036.57 & 7400.79 & 96.4098 \\
    \rowcolor{black!8} {SceneCode} & \textbf{1.5738} & \textbf{0.6066} & \textbf{0.0000} & \textbf{6013.64} & \textbf{4945.07} & \textbf{22.0328} \\
    \bottomrule
  \end{tabular}
  }
\end{table}

\begin{figure}[t]
  \centering
  \includegraphics[width=\linewidth]{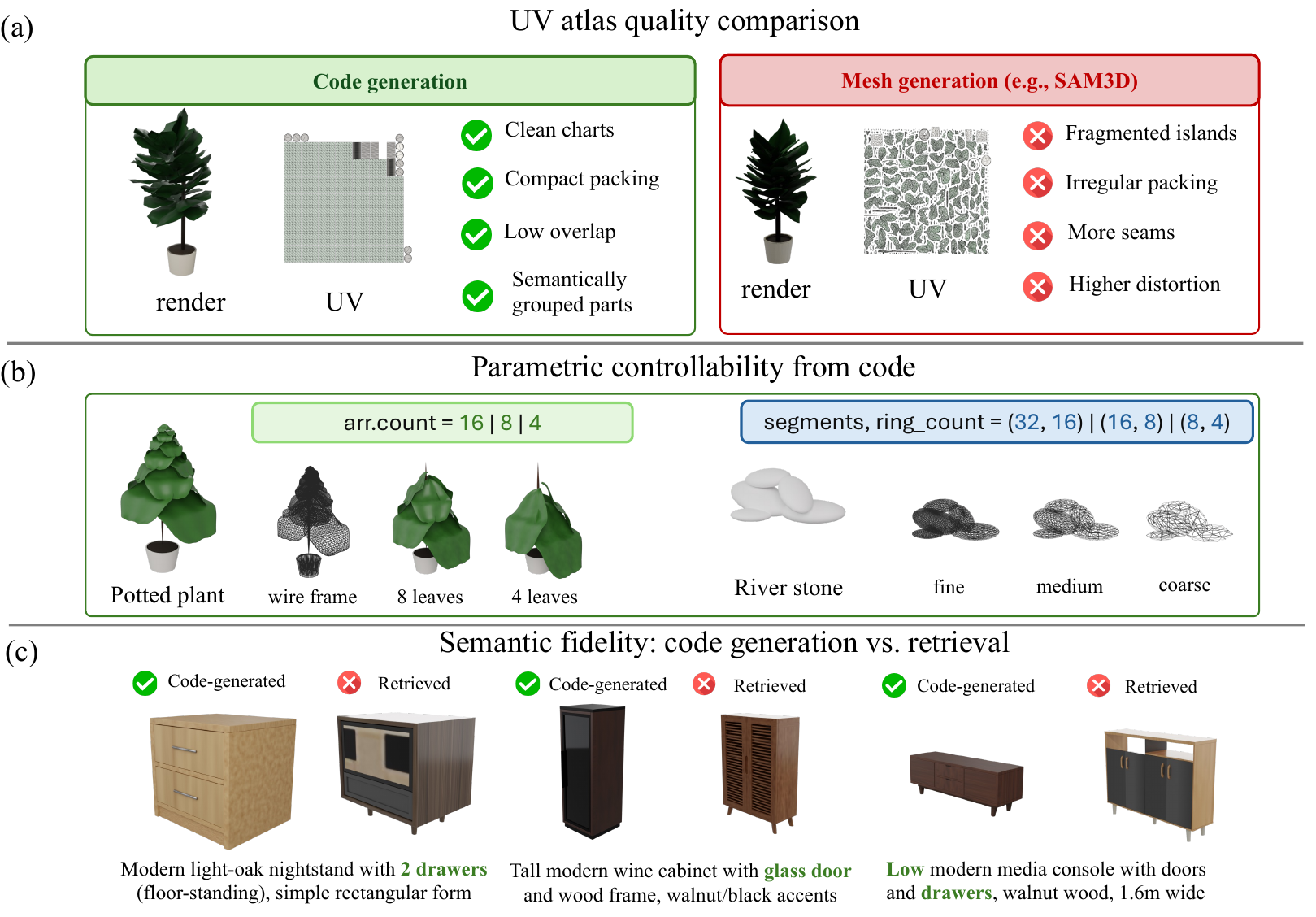}
  \caption{\textbf{Object-level qualitative comparison.} (a) Mesh and UV layouts of representative assets from {SceneCode} versus SAM 3D Objects. (b) Code-level editability: locally re-executing one object program with different parameters yields variant objects. (c) On-demand articulated objects with prescribed structure or material that retrieval-based pipelines cannot satisfy.}
  \label{fig:object_level_results}
\end{figure}

\paragraph{Mesh and UV usability.}
Table~\ref{tab:object_level_results} shows that {SceneCode} dominates SAM 3D Objects on every metric designed to capture editability and simulation-oriented usability.
The largest gap is on UV organization: {SceneCode} produces about 22 UV islands per asset on average, roughly 4.4$\times$ fewer than the 96 islands of SAM 3D Objects, which means a far more coherent UV parameterization for downstream texturing and editing.
Geometric compactness improves in parallel: face count drops to 6013.64 from 12036.57 and vertex count to 4945 from 7400, roughly halving storage and rendering cost while still expressing the requested object.
Besides, {SceneCode} averages 1.5738 material slots per asset and a PBR channel coverage of 0.6066, indicating more complete material modeling.
Topology is also cleaner: {SceneCode} produces 0 non-manifold edges on average, whereas the image-to-3D pipeline still generates assets with non-manifold artifacts that downstream simulators would need to repair.
This is a direct consequence of the part-wise primitive program: each part owns its own mesh, materials, and UV layout, so the asset never has to be flattened into a single watertight blob.
Figure~\ref{fig:object_level_results}(a) makes the gap visible: the SAM 3D Objects potted plant show fragmented UV charts, while the {SceneCode} counterparts have a small set of large, regularly shaped islands that align with the semantic parts of the object.

\paragraph{Code-level editability.}
A second advantage of representing objects as programs is that they remain editable after generation.
Because each part is constructed by primitive-level Blender Python code with explicitly named parameters, individual attributes can be modified and re-executed locally without re-running the room-level agent or regenerating the reference image.
Figure~\ref{fig:object_level_results}(b) illustrates this: starting from a single potted-plant program, we vary parameters such as \texttt{arr\_count} (the number of leaf instances) and \texttt{ring\_count} (the angular and radial subdivisions of each leaf surface) to produce coarse, medium, and fine variants of the same object identity.
This kind of structured, parameter-level edit is difficult to obtain from an image-to-3D mesh, where the asset is delivered as a single opaque triangle soup with no semantic handles to address.

\paragraph{On-demand articulated objects.}
Finally, code-driven assets remove the dataset bottleneck for articulated objects.
Retrieval-based pipelines, including the articulated branches of recent simulation-ready scene generators, can only return objects that already exist in a curated articulated-object library, so requests for unusual functional structures (e.g., a cabinet with a specific number of drawers) or unusual materials (e.g., a glass-fronted cabinet) typically have to be approximated by the closest available asset.
Figure~\ref{fig:object_level_results}(c) shows two such requests where the agent specifies the functional structure or material of the target object.
Because {SceneCode} compiles each request into a part-decomposed program with explicit movable parts, it can synthesize a customized articulated asset that exactly matches the request, including the prescribed part count and material, rather than collapsing the request onto whatever happens to be available in the dataset.
This point is detailed in Appendix~\ref{app:scenecode_vs_baselines}.

\subsection{Articulation and Robot Interaction}
\label{sec:exp_articulation}

\begin{figure}[ht]
  \centering
  \includegraphics[width=\linewidth]{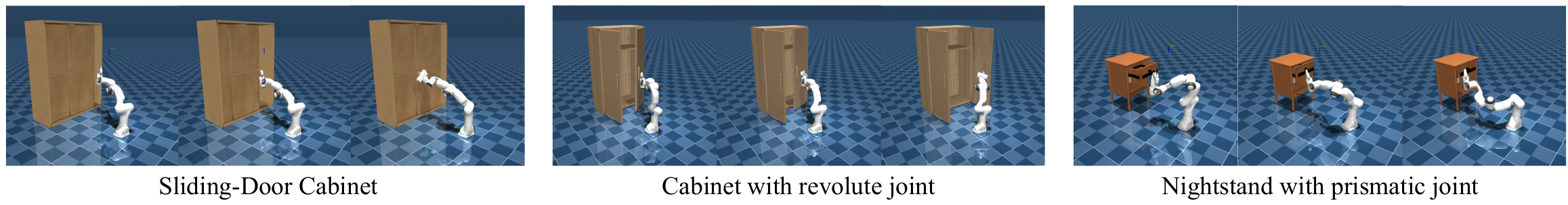}
  \caption{\textbf{Robot interaction with generated articulated assets.}
  A SceneCode-generated articulated object is imported into MuJoCo for contact-based robot manipulation. The movable parts produced by the Articulated Object Program remain independent links with compiled joints, allowing the robot to physically open or slide them.}
  \label{fig:mujoco_interaction}
\end{figure}

Figure~\ref{fig:mujoco_interaction} demonstrates that objects generated by our Articulated Object Program are not only visually plausible but also physically interactive. Since movable components are preserved as separate links and exported with joint metadata, the resulting assets can be loaded into MuJoCo~\citep{todorov2012mujoco} and manipulated by a robot through contact. This shows that SceneCode produces articulated objects suitable for downstream simulation and embodied interaction.

\section{Conclusion}
\label{sec:conclusion}

We presented \textsc{SceneCode}, a code-driven framework for physically interactable indoor scene synthesis. By representing objects as executable, part-aware programs rather than opaque meshes, SceneCode couples visual geometry, semantic structure, materials, articulation, and simulation metadata in a unified form. Combined with a room-level agentic planner and execution-guided object program synthesis, the system produces prompt-faithful scenes with editable and interaction-ready assets. Experiments show advantages in semantic fidelity, object-level mesh usability, human preference, and downstream robot interaction. 


\medskip
{
\small
\bibliographystyle{plainnat}
\bibliography{neurips_2026}
}

\newpage
\appendix
\section*{Appendix}

{\large\bfseries Contents}

\vspace{0.75em}
\noindent\begin{tabular*}{\textwidth}{@{}p{0.86\textwidth}@{\extracolsep{\fill}}r@{}}
A. More Related Work & \pageref{app:more_related_work} \\
B. Route-Specific Prompt Design for Code-Driven Object Generation & \pageref{app:prompt_design} \\
C. Evaluation Prompts & \pageref{app:evaluation_prompts} \\
D. Metric Details & \pageref{app:metric_details} \\
E. SceneCode Assets versus Prior Asset Strategies & \pageref{app:scenecode_vs_baselines} \\
F. More Analysis & \pageref{app:more_analysis} \\
G. Full User Study Results & \pageref{app:user_study} \\
H. Computational Cost Statistics & \pageref{app:cost_statistics} \\
I. Additional Qualitative Demonstrations & \pageref{app:more_demos} \\
J. Executable Object-Code Demonstration & \pageref{app:code} \\
K. Limitations & \pageref{app:limitations} \\
\end{tabular*}

\vspace{2em}

\section{More Related Work}
\label{app:more_related_work}

\paragraph{Additional Indoor Scene Synthesis Systems.}
Beyond the representative works cited in the main text, other learning-based synthesizers include SceneFormer~\citep{wang2021sceneformer}, while a broader set of LLM- and retrieval-guided pipelines populate rooms by querying curated 3D libraries, including LayoutGPT~\citep{feng2023layoutgpt}, AnyHome~\citep{wen2023anyhome}, InstructScene~\citep{lin2024instructscene}, and Architect~\citep{wang2024architect}.
Procedural systems such as ProcTHOR~\citep{deitke2022procthor} scale environment diversity with hand-crafted rules.
These systems differ in how they plan layouts but commonly rely on retrieved static meshes, which is the dependency our code-driven asset layer removes.

\paragraph{Additional Code-Driven 3D Generation Systems.}
Beyond the works cited in the main text, recent VLM-driven code-generation systems include 3D-GPT~\citep{sun20233dgpt} and BlenderLLM~\citep{huang2024blenderllm}, which synthesize Blender Python or domain-specific code from natural language for either single-object modeling or scene-level visual layout.

\paragraph{Articulated Object Generation.}
Articulated assets used in embodied AI predominantly come from curated datasets such as PartNet~\citep{mo2019partnet} and PartNet-Mobility / SAPIEN~\citep{xiang2020sapien}, which support manipulation learning~\citep{wang2022adaafford} but cap the categories and joint configurations available downstream.
A growing line of work generates articulated objects directly: NAP~\citep{lei2023nap} and CAGE~\citep{liu2024cage} learn distributions over part graphs and joint parameters, while URDFormer~\citep{chen2024urdformer} infers URDFs from real-world images.
Methods such as NAP, CAGE, and URDFormer focus on articulated object priors or URDF inference, whereas SceneCode focuses on generating full indoor scenes whose assets are represented as executable programs and registered in a persistent world state.
Thus, our goal is not to directly benchmark articulated-object correctness against these methods, but to address scene-level integration and executable representation for interactable indoor environments.

\section{Route-Specific Prompt Design for Code-Driven Object Generation}
\label{app:prompt_design}
\label{app:prompts}

This appendix details the route-specific prompting design used by the code-driven object generation module described in Section~\ref{sec:object_program}. Although every routed object follows the same downstream pipeline---asset request routing, ObjectPlan construction, ObjectPlan verification, part-wise Blender program synthesis, execution-guided repair, and asset packaging---different indoor object families benefit from different geometric construction priors. Wall art is best represented as a thin textured canvas; large furniture requires stable support surfaces and repeated structural components; simple manipulands should avoid unnecessary part decomposition; and articulated objects must preserve movable parts as independent links for downstream joint compilation.

To accommodate these differences without fragmenting the pipeline, we adopt a small set of specialized prompt profiles. All profiles share a common ObjectPlan schema and the same planner--checker--constructor structure, but each profile augments the shared base with route-specific construction priors that guide the VLM toward executable and geometrically reliable Blender Python code. The prompt features used in our implementation are summarized in Table~\ref{tab:prompt_feature_summary}, and the mapping from generation strategies to feature codes and target object categories is given in Table~\ref{tab:strategy_prompt_features}.

\begin{table}[t]
    \centering
    \small
    \renewcommand{\arraystretch}{1.18}
    \begin{tabular}{p{0.08\linewidth} p{0.84\linewidth}}
    \toprule
    \textbf{Code} & \textbf{Prompt feature} \\
    \midrule
    Base 
    & Shared ObjectPlan schema. Each generated object is represented by a structured plan containing object category, semantic parts, part dimensions, local position, rotation, priority, and material. The planner, checker, and constructor agents use this plan to synthesize Blender Python code. \\
    
    A 
    & Explicit mirror reference. Mirror operations are required to use an explicit symmetry reference object located at the intended symmetry origin, instead of relying on the default object origin. This avoids incorrect mirroring around each part's local origin. \\
    
    B 
    & Explicit pivot for radial copies. For rotational arrays or radial copies, both the object origin and the Empty pivot are placed at the desired rotation center. This avoids spiral-like artifacts or accumulated translation errors during repeated rotation. \\
    
    C 
    & Controlled curve construction. Free-form Bezier curves are avoided for geometry-critical shapes. Curves and arcs are instead constructed from mathematically sampled points using \texttt{POLY} splines or analytic primitives. \\
    
    D 
    & Boolean topology constraint. When Boolean cuts produce N-gon faces, subdivision modifiers are disabled to avoid unstable topology and rendering artifacts. When smoothing is needed, bevel operations are preferred. \\
    
    E 
    & Few-shot demos for high-risk geometry. Prompt profiles include construction examples for geometries that are difficult for VLMs to generate reliably, such as hollow bowls and cups using shell-first modeling with Solidify, parameterized fork/spoon meshes, concave shapes, and custom curved surfaces. \\
    
    F 
    & Explicit internal structure. Objects with meaningful interiors, such as cabinets, desks, drawers, shelves, and dishwashers, are required to include \texttt{sub\_parts} describing inner walls, shelves, drawer bottoms, rails, or racks rather than modeling only the exterior shell. \\
    
    G 
    & Reuse for repeated structures. Repeated homogeneous components, such as drawers, doors, handles, legs, rails, shelves, and slats, are generated from a prototype function and then duplicated with independent geometry using \texttt{copy()} and \texttt{data.copy()}. This improves consistency and reduces code complexity. \\
    
    H 
    & Material templates. Prompts provide standard material construction templates using Principled BSDF, procedural textures, and material reuse. Some profiles additionally require explicit sRGB-to-linear conversion for image or color inputs. \\
    
    I 
    & Blender engineering rules. Generated scripts follow export-friendly Blender conventions: avoid relying on parent hierarchies for final assets, use absolute coordinates in the object-local frame, apply scale and rotation before final placement, clean same-name objects before construction, and use standardized object, mesh, material, and collection names. \\
    
    J 
    & Movable-part independence and articulation metadata. Movable parts must be marked with \texttt{is\_movable=true} and \texttt{must\_be\_independent=true}. They must not be fused through Array or Mirror operations. The articulated branch additionally generates URDF/SDF link and joint metadata, including pivot, axis, joint type, and motion limit. \\
    
    K 
    & Wall-art constraints. Wall-art objects must include a \texttt{canvas}, avoid unnecessary \texttt{glass\_cover} geometry, use a thin depth, orient the front face toward \texttt{-Y}, apply a local cropped image as the canvas material, and explicitly define UV coordinates for the front face. \\
    
    L 
    & Simple-object constraints. Simple manipulands should use a small number of parts and avoid over-decomposition. Bowl-, cup-, and plate-like objects are encouraged to use shell construction with Solidify to obtain physically meaningful thickness. \\
    
    M 
    & Route-level constraints. The routing prompt preserves the original requested style and avoids inventing unsupported materials or colors. Composite requests are decomposed into separate assets when appropriate, and objects outside the current agent's category are filtered to prevent duplicated or misplaced generation. \\
    \bottomrule
    \end{tabular}
    \caption{Prompt feature taxonomy used by the route-specific code generation profiles. All routes share the Base ObjectPlan structure, while additional constraints provide object-family-specific construction priors.}
    \label{tab:prompt_feature_summary}
    \end{table}
\begin{table}[t]
    \centering
    \small
    \renewcommand{\arraystretch}{1.18}
    \begin{tabular}{p{0.20\linewidth} p{0.22\linewidth} p{0.50\linewidth}}
    \toprule
    \textbf{Generation strategy} & \textbf{Feature codes} & \textbf{Corresponding objects} \\
    \midrule
    \textbf{Wall-Art Program} 
    & Base, A, G, H, I, K, M 
    & Print-like or image-based wall-mounted objects, including artwork, paintings, posters, prints, canvas prints, murals, framed artwork, framed photos, and picture frames. These objects are represented as thin canvases or frames with explicit UVs and image materials. \\
    
    \textbf{Static Furniture Program} 
    & Base, A, B, E, F, G, H, I, M 
    & Large rigid furniture and default rigid objects without functional moving parts, including plain tables, chairs, stools, benches, sofas, beds, open shelves, ladder shelves, plant stands, straight floor lamps, ladders, and vehicle-like rigid props. This branch also handles non-art wall-mounted rigid objects such as mirrors, open wall shelves, mounted TVs, sconces, light switches, and ceiling fixtures when no functional articulation is required. \\
    
    \textbf{Simple Manipuland Program} 
    & Base, A, C, D, E, H, I, L, M 
    & Structurally simple rigid objects with a single dominant shape, thin or flat form, or straightforward silhouette, including bowls, forks, spoons, vases, dishes, plates, rigid coasters, books, whole fruits, tape, scissors, and saucers. These objects use compact primitive or custom-mesh programs and avoid unnecessary part decomposition. \\
    
    \textbf{Structured Manipuland Program} 
    & Base, A, C, D, E, G, H, I, J$^\ast$, M 
    & Rigid objects with moderate structural complexity but no functional articulation, including cups, mugs, chopsticks, hardware, simple table lamps, standalone lids, unmounted rigid TVs, USB sticks, remote controls, cables, phones, toys, crates, tires, flower pots, and potted plants. This branch preserves visible subparts such as handles, shells, housings, cables, and repeated components while packaging the object as a rigid asset. \\
    
    \textbf{Articulated Object Program} 
    & Base, A, B, E, F, G, H, I, J, M 
    & Objects with functional movable parts. Furniture examples include drawers, hinged doors, sliding doors, dressers, nightstands, filing cabinets, wardrobes, cabinets, refrigerators, freezers, ovens, microwaves, dishwashers, washing machines, dryers, office chairs, and desks with drawers. Manipuland examples include bottle caps, trash-can lids, monitor tilt mechanisms, hinged boxes, bucket handles, cameras, eyeglass arms, rotating globes, toaster levers, kettle lids, thermos caps, and laptop hinges. Wall-mounted articulated examples include wall clocks, pendulum or cuckoo clocks, and wall-mounted mailboxes. This branch additionally compiles movable parts into URDF/SDF links and prismatic or revolute joints. \\
    \bottomrule
    \end{tabular}
    \caption{Route-specific generation strategies. Each strategy uses a specialized prompt profile while sharing the same downstream ObjectPlan verification, part-wise code synthesis, execution-guided repair, and asset packaging pipeline. $J^\ast$ indicates that the structured manipuland prompt preserves independence for potentially movable-looking subparts, but the object is treated as a rigid asset unless routed to the articulated branch.}
    \label{tab:strategy_prompt_features}
    \end{table}

\subsection{Routing Strategies and Object Categories}
\label{app:routing_strategies}

The asset router maps each room-level \texttt{AssetRequest} to one of five generation strategies. The decision is made jointly from the requested object category, textual description, support relation, expected size, and functional affordances, and follows a fixed priority order. Requests with functional movable components---such as doors, drawers, lids, hinges, sliders, or rotating parts---are routed to the \emph{articulated} branch first, since these components must be preserved before any geometry export. Print-like wall-mounted objects are routed to the \emph{wall-art} branch. Among the remaining rigid objects, large furniture-scale items without functional articulation are routed to the \emph{static furniture} branch, while tabletop or handheld objects are routed to either the \emph{simple manipuland} or the \emph{structured manipuland} profile according to their expected part complexity.

It is important to emphasize that routing does not change the downstream representation: every generated asset is still produced by code-generated geometry and registered as a scene object through the same packaging stage. Routing only selects an object-family-specific prompt profile that supplies the appropriate geometric and functional priors for VLM-based program synthesis.

\subsection{Prompt Feature Taxonomy and Targeted Failure Modes}
\label{app:prompt_constraints}

The route-specific features in Table~\ref{tab:prompt_feature_summary} were introduced to address recurring failure modes that we observed when prompting a VLM with a single universal prompt. Such a universal prompt frequently produces plausible-looking but geometrically fragile code, particularly for symmetry operations, radial copies, concave structures, hollow objects, and objects with repeated internal components. For instance, mirror operations may be incorrectly performed around each part's local origin instead of the desired symmetry plane (addressed by feature~A); radial copies may drift when the pivot is not explicitly placed at the rotation center (feature~B); free-form B\'ezier curves are difficult to control when exact arcs or repeated curved elements are required (feature~C); and applying subdivision modifiers to Boolean-cut N-gon faces tends to introduce unstable topology and rendering artifacts (feature~D).

To improve robustness, each prompt profile encodes only the construction priors that are relevant to its routed object family. Static furniture and articulated furniture profiles emphasize part decomposition, internal structure, repeated components, and support surfaces (features~F and~G). Simple manipulands emphasize compact geometry programs and explicitly discourage over-decomposition (feature~L). Structured manipulands emphasize visible subparts while still packaging the object as a rigid asset (feature~J$^\ast$). Wall-art prompts emphasize canvas construction, UV mapping, and image material assignment (feature~K). Articulated prompts additionally require movable parts to remain independent so that they can be converted into URDF/SDF links and joints during downstream compilation (feature~J). Several engineering-level rules---few-shot demonstrations for high-risk geometry (E), standardized material templates (H), export-friendly Blender conventions (I), and route-level style preservation (M)---are shared across most profiles.

These constraints are not intended to replace execution-time validation. Instead, they reduce the search space for the VLM before code generation. The generated ObjectPlan is still verified before construction, and the resulting Blender programs are executed in a headless environment with up to three repair attempts when runtime errors occur, followed by up to two critic-driven refinement iterations on the rendered output.

\subsection{Interaction with Articulated Asset Compilation}
\label{app:articulated_routing}

The articulated route is given priority over the rigid-object routes whenever the asset request implies a functional moving part, because such components must be preserved before geometry export. A nightstand with drawers, a cabinet with doors, or a box with a hinged lid cannot be generated as a single fused mesh if it is expected to support physical interaction. The articulated prompt therefore marks movable components with \texttt{is\_movable=true} and \texttt{must\_be\_independent=true}, prevents them from being merged through Array or Mirror operations, and records the auxiliary information required for joint compilation.

After part-wise code generation, articulated objects are converted into a link--joint representation. Each movable part is assigned a parent link, child link, joint type, joint axis, pivot or origin, and motion limit. We currently support revolute joints for hinged components and prismatic joints for sliding components. The resulting articulated asset is exported to URDF and converted to SDF for downstream simulation. In contrast, objects routed to the simple manipuland, structured manipuland, static furniture, or wall-art profiles are packaged as rigid assets unless the router explicitly selects the articulated branch.

\subsection{Shared Structure and Style Preservation Across Profiles}
\label{app:prompt_profiles}

All prompt profiles share the same high-level planner--checker--constructor structure: the planner produces an ObjectPlan, the checker revises the plan for completeness and geometric plausibility, and the constructor synthesizes Blender Python programs for individual parts. The differences between profiles lie entirely in their construction priors. The wall-art profile requires an explicit canvas and UV-mapped image texture; the simple-object profile discourages unnecessary part splitting; the furniture profiles encourage repeated structural components and meaningful internal subparts; and the articulated profile requires independent movable links and downstream joint metadata.

The routing prompt itself is deliberately conservative. It preserves the user-requested style and avoids inventing unsupported materials, colors, or functional mechanisms unless they are implied by the asset request. When a request contains multiple distinct objects, the router may split it into separate asset requests so that each object can be generated by the appropriate profile. This prevents, for example, a framed poster and a shelf from being incorrectly generated by a single construction strategy.

Overall, the route-specific prompt design improves the reliability of VLM-based Blender code generation by aligning the prompt constraints with the geometric and functional structure of each object family, while keeping the downstream verification, execution, repair, and packaging pipeline unified across all routes.

\subsection{Why Route-Specific Construction Prompts Matter}
\label{app:route_specific_prompt_ablation}

\begin{center}
  \centering
  \includegraphics[width=\linewidth]{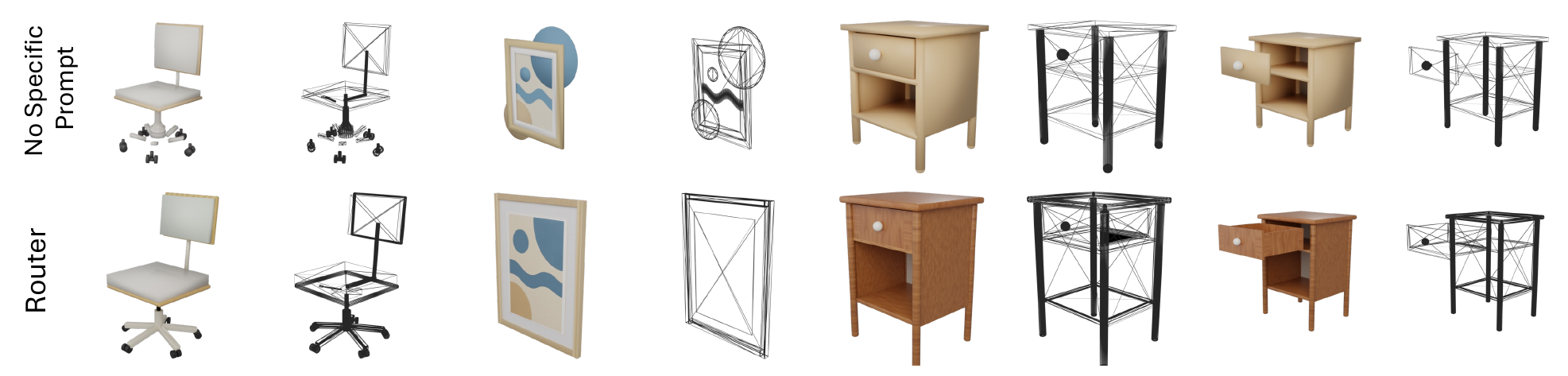}
  \captionof{figure}{\textbf{Route-specific prompt ablation.}
  Generic prompting can produce visually plausible assets, but route-specific construction prompts better preserve semantic parts, internal structure, material grounding, and articulation-ready components.}
  \label{fig:route_specific_prompt_ablation}
\end{center}

Here, \emph{generic prompting} refers to removing the router and simply asking the VLM, given the input request, to generate Blender code for the desired 3D content. As shown in Figure~\ref{fig:route_specific_prompt_ablation}, this baseline can produce visually plausible and metric-friendly meshes, but it often fails to preserve semantic parts, internal structure, material grounding, and articulation readiness. For example, the generic nightstand lacks internal structure and drawers, making it unsuitable for downstream articulation. More complex geometry, such as office-chair structure, and more complex material grounding, such as canvas construction and texture assignment, also remain difficult without route-specific construction priors.

\section{Evaluation Prompts}
\label{app:evaluation_prompts}

This appendix lists the complete natural-language prompt descriptions used in our
scene-level evaluation benchmark. The prompt set contains 30 indoor scene
descriptions drawn from six room categories: bedroom, living room, dining room,
kitchen, bathroom, and basement. The category grouping below follows the room
type stated in each prompt description. The full benchmark CSV also records a
\texttt{Difficulty} label per prompt; in total there are 15 easy prompts,
12 medium prompts, and 3 hard prompts.

\subsection{Bedroom}
\label{app:eval_prompts_bedroom}

\begin{itemize}
    \item A bedroom with a bed and a wardrobe.
    \item A bedroom with a king-size bed in the corner of the room, two large blue armchairs, and a floor lamp near a armchair.
    \item A bedroom with a double bed and a mini fridge near the bed, a table across from the door, and a painting on the wall above the bed.
    \item A bedroom with a king-size bed positioned against the wall across from the window. Two stools are placed in front of the bed. A painting hangs above the bed, flanked by a wall light on each side. There is a bookcase with two books on the left side.
    \item A teenager's bedroom features a comfortable twin bed with a backboard in the far corner, with boxes underneath it. At the foot of the bed is a small desk equipped with a monitor, an external keyboard and mouse, and a desk lamp on the right for visibility, accompanied by a rolling chair. Next to the bed, a nightstand with an additional floor lamp nearby provides space for a phone and other valuables. A sizable wooden wardrobe with multiple drawers offers ample storage for clothes, while a coffee table beside it holds books and board games. In the center of the room, a tan-colored rug creates a cozy spot to sit, and the walls are adorned with various posters and pictures.
    \item This is a small, plain bedroom. Upon entering through the door, a white desk with a black rolling chair in front of it is positioned against the wall to the right. A queen bed is centered against a wall, with two modern nightstands beside it, one of which holds a stylish lamp.
\end{itemize}

\subsection{Living Room}
\label{app:eval_prompts_living_room}

\begin{itemize}
    \item A living room with a TV, sofa, bookshelf, and coffee table.
    \item A living room with a two-seater sofa against the wall, a square rug in the middle in front of the sofa, and two large plants on the floor near the sofa.
    \item A large living room with three display shelves against the wall, with a long sofa in front. There are two tables in front of the sofa and another table right of the sofa with a lamp on top. There is no TV in the room.
    \item A living room featuring an irregular-shaped table in the middle of the room with a sofa positioned in front of it. Across the table are two sofa chairs with a small wooden coffee table placed between them. A clock is mounted on the wall far from the door.
    \item The room is dimly lit, creating a somber atmosphere in a cozy and well-furnished living room. In the center of the room, there is a dining table with four wooden dining chairs arranged neatly around it. The table appears to be set, indicating a recent meal or gathering. Adjacent to the dining area, There is a glass coffee table serving as a centerpiece for the seating arrangement. It is both stylish and practical, providing a surface for drinks or decor. A multi-seat sofa is in front of the coffee table, providing ample seating for guests. Completing the seating options, two comfortable armchairs also face the coffee table on opposite sides near the sofa. Next to the sofa and each armchair is a corner side table, adding a touch of functionality and convenience. Each table has a lamp on it for lighting. The sofa and seating arrangement should face an opposite wall, against which someone in the scene could view a large flatscreen TV. The TV is supported by a TV stand with storage underneath for various books. The room appears to reflect the inhabitants' taste for a warm and inviting environment, despite the unsettling news program playing on the TV in the background. The juxtaposition of the serene living room with the chaos and screams on the TV screen creates a sense of tension and unease.
    \item A living room featuring an overstuffed sofa, a vintage wooden table, alongside a bicycle hung decoratively on the wall.
\end{itemize}

\subsection{Dining Room}
\label{app:eval_prompts_dining_room}

\begin{itemize}
    \item A dining room with two bar stools at the short sides of a bar table.
    \item A dining room with a cabinet next to the door against the wall, and two wine cabinets against the wall near a bar table.
    \item A dining room with a table in the corner of the room and a chair on the long side.
    \item A dining room with a bar table positioned in the middle of the room. A wooden shelf is mounted on the wall, holding a potted plant and a jug, adding a touch of decor and functionality. Along the wall, there is a fridge and a wine cabinet, providing ample storage and a cohesive design.
    \item A dining room with six wooden dining chairs surrounding a round wooden table in the middle of the room. There is no coffee table in the room.
    \item A dining room with a circular table surrounded by six vintage chairs, and an old wooden ladder against the wall displaying plants and decorative jars.
\end{itemize}

\subsection{Kitchen}
\label{app:eval_prompts_kitchen}

\begin{itemize}
    \item A kitchen featuring a bistro table with two chairs in a corner and a refrigerator positioned nearby.
    \item A stylish kitchen with an elegant pendant light hanging over a marble-topped kitchen counter. A single cabinet is positioned against the wall, providing ample storage space.
    \item In this modern kitchen, a large bar table with four high-backed stools stands in the center, facilitating social gatherings, while a stainless steel fridge is positioned against the far wall, next to a sleek pantry shelf. A small bowl of fruit sits on the table's surface.
    \item This cozy kitchen features a rustic wooden dining table positioned against the window, surrounded by four chairs. A sideboard against the wall is equipped with a coffee maker and a small herb planter.
\end{itemize}

\subsection{Bathroom}
\label{app:eval_prompts_bathroom}

\begin{itemize}
    \item A bathroom with a freestanding bathtub positioned beside a frosted window and a double vanity located across from it.
    \item In a classic bathroom, a double-sink vanity is set against one wall with two round mirrors hanging above, while a toilet is positioned in the corner.
    \item In this bathroom, a wide vanity sits against the wall, paired with a tall mirror mounted above. Nearby, a stylish laundry hamper is positioned next to the door, while a small basket containing bath essentials rests beside the tub.
    \item A bright bathroom features a spacious bathtub positioned beside a large window, while a stylish shelving unit stands against the wall, displaying towels and small decorative items. The toilet is placed across from the tub, with a small basket beside it.
\end{itemize}

\subsection{Basement}
\label{app:eval_prompts_basement}

\begin{itemize}
    \item In this basement, a pool table occupies the center of the room, with two bar stools lined up at a nearby bar table, allowing for social gatherings and entertainment.
    \item A gym-purpose basement featuring a treadmill and an exercise bike positioned against one wall, with a tall mirror mounted on a wall directly facing the exercise bike.
    \item This versatile basement features a large sofa and an area rug in the center, creating a cozy movie area, while a small treadmill is positioned in the corner for quick exercise sessions. A mini chalkboard hangs on the wall next to a wall clock.
    \item This entertaining basement layout features a large gaming setup with two monitors on a desk against one wall, while a comfy bean bag chair is positioned nearby for casual seating. Across from the gaming area, a small cabinet with a mini fridge and a popcorn machine on top completes the setup.
\end{itemize}

\section{Metric Details}
\label{app:metric_details}

\subsection{Scene-Level Metrics}
\label{app:scene_level_metrics}

We use the following scene-level metrics from the SceneEval benchmark~\citep{tam2026sceneeval}, which jointly assess explicit prompt requirements (object presence, attributes, and relations) and implicit physical plausibility (support, accessibility, navigability, collision, out-of-bound placement, and opening clearance).

\paragraph{CNT (Object Count).}
Satisfaction rate for object count requirements.
A VLM maps scene objects to objects specified in the prompt, then compares instance counts to annotated quantities (exact or relative).
This metric measures whether the generated scene contains the requested number of objects per category.

\paragraph{ATR (Object Attribute).}
Satisfaction rate for object attributes (e.g., colors, materials).
For each object with attribute requirements, a front view and a reference view with a human figure for scale are rendered, and a VLM evaluates whether the object satisfies the specified attributes.
This metric checks whether the generator preserves descriptive prompt modifiers in addition to producing the correct object categories.

\paragraph{OOR (Object--Object Relationship).}
Satisfaction rate for spatial relationships between objects (e.g., ``chair in front of desk'').
Relations are evaluated against a predefined set of relationship types using geometric checks on object poses and bounding boxes.
This metric measures whether requested inter-object spatial constraints are realized in the generated scene.

\paragraph{OAR (Object--Architecture Relationship).}
Satisfaction rate for object--architecture relationships (e.g., ``sofa against wall'', ``rug in middle of room'').
Relations between objects and architectural elements (walls, floor, ceiling) are checked against a predefined set of relationship types.
This metric measures whether requested object-to-architecture spatial constraints are satisfied.

\paragraph{SUP (Support).}
Fraction of objects that are stably supported.
Each object is classified into one of four support types --- ground, object, wall, or ceiling --- by a VLM from rendered images, and ray casting verifies whether the object is actually supported by other objects or architectural elements consistent with that type.
This metric captures basic placement stability beyond non-collision.

\paragraph{ACC (Accessibility).}
Fraction of objects whose functional sides remain accessible.
A VLM identifies the functional sides of each object (e.g., the front of a sofa, the sides of a bed), then a 2D occupancy analysis checks whether those sides are unobstructed.
This metric measures whether placements preserve usability of the generated objects.

\paragraph{NAV (Navigability).}
Ratio of the largest connected free floor region to the total free floor space.
We construct a 2D occupancy projection of the scene and compute connected components on the free space.
Lower values indicate that object placements fragment the floor into isolated regions, while values close to one indicate a single, navigable free region.

\paragraph{COL (Collision).}
Fraction of objects that intersect another object.
We perform pairwise mesh-based collision tests between all objects and report the percentage of objects participating in at least one collision.
This metric captures the most basic physical-validity requirement: objects in a scene should not overlap.

\paragraph{OOB (Out of Bounds).}
Fraction of objects placed outside the floor plan boundary.
For each object, surface points are sampled and rays are cast toward the floor; an object is considered out-of-bounds if fewer than 99\% of its rays hit the floor of the room.
This metric penalizes placements that protrude through walls or land outside the room footprint.

\paragraph{OPC (Opening Clearance).}
Fraction of architectural openings (doors and windows) whose clearance volume is obstructed by objects.
For each opening, an extruded clearance box is constructed in front of the opening, and any object intersecting this volume is treated as blocking the opening.
This metric ensures that doors and windows remain functionally usable.

\subsection{Object-Level Mesh and Material Metrics}
\label{app:object_level_metrics}

We evaluate generated 3D assets using mesh- and material-level metrics that reflect downstream usability.
Specifically, we report material slot count, PBR channel coverage, non-manifold edge count, total face count, total vertex count, and UV island count.
These metrics capture complementary aspects of asset quality, including material editability, PBR rendering readiness, topological validity, geometric compactness, and UV layout organization.
All object-level metrics are computed with the same asset loading and mesh evaluation pipeline, first per generated asset and then averaged over the evaluation set.

\paragraph{Material slot count.}
For each generated asset, we load all mesh objects belonging to the asset and count the non-empty material slots assigned to them.
For a multi-part asset, the asset-level score is the sum of material slots over its mesh parts.
At a high level, this metric measures material editability and semantic material decomposition: a larger number of slots indicates that the asset is divided into more material regions, which is often useful for editing and assigning part-specific appearance.

\paragraph{PBR channel coverage.}
We evaluate PBR readiness over a fixed set of material channels $\mathcal{C}_{\mathrm{PBR}}$, such as base color, roughness, metallic, normal, alpha, and emission channels.
For each material $m$, we count how many channels in $\mathcal{C}_{\mathrm{PBR}}$ are present as assigned scalar values or texture/node inputs, and normalize by the number of considered channels:
\[
S_{\mathrm{pbr}}(m)=
\frac{|\{c\in\mathcal{C}_{\mathrm{PBR}}: c\ \mathrm{is\ present\ in}\ m\}|}
{|\mathcal{C}_{\mathrm{PBR}}|}.
\]
The asset-level score is the average over its materials, and the reported score is averaged over all generated assets.
The purpose of this metric is to measure whether an asset is ready for physically based rendering workflows, where richer channel coverage gives downstream renderers and editors more complete material information.

\paragraph{Non-manifold edge count.}
For each mesh, we count edges whose incident-face count is not equal to two, including boundary edges and edges shared by more than two faces.
For an asset $a$ with mesh parts $\mathcal{M}_a$, the asset-level non-manifold count is
\[
N_{\mathrm{nonmanifold}}(a)=
\sum_{M\in\mathcal{M}_a}
|\{e\in E_M: \deg_F(e)\neq 2\}|,
\]
where $\deg_F(e)$ is the number of faces incident to edge $e$.
This metric measures topological validity and mesh cleanliness.
Lower values indicate cleaner geometry, and zero means that no non-manifold edges are detected by the evaluation pipeline.

\paragraph{Total face count.}
Total face count is computed by summing the number of mesh faces over all mesh parts in the asset:
\[
N_{\mathrm{face}}(a)=\sum_{M\in\mathcal{M}_a}|F_M|.
\]
Its high-level purpose is to measure geometric compactness and downstream processing cost.
Under comparable visual quality, a lower face count indicates a more efficient representation that is cheaper to render, store, and process.

\paragraph{Total vertex count.}
Total vertex count is computed by summing the number of mesh vertices over all mesh parts in the asset:
\[
N_{\mathrm{vertex}}(a)=\sum_{M\in\mathcal{M}_a}|V_M|.
\]
Like face count, this metric measures geometric compactness.
Lower values suggest a lighter mesh representation, assuming the generated asset preserves similar visual and semantic quality.

\paragraph{UV island count.}
For each mesh with a UV map, we build the UV connectivity graph and count connected components in the UV layout.
Faces are connected in this graph when their corresponding UV edges are continuous in UV space.
The asset-level UV island count is the sum of UV connected components over all mesh parts.
This metric measures UV layout organization: fewer islands indicate a more coherent parameterization that is easier to inspect, texture, and edit, while excessive fragmentation suggests a less organized UV layout.

\section{SceneCode Assets versus Prior Asset Strategies}
\label{app:scenecode_vs_baselines}

This appendix clarifies that \textsc{SceneCode} is not merely changing the source from which assets are obtained. Instead, it changes the representation of the asset layer. Prior pipelines typically retrieve a fixed mesh, retrieve an already-articulated model, or generate an opaque mesh. In all three cases, the asset is largely treated as an endpoint. \textsc{SceneCode} instead represents an asset as executable, part-aware construction code that can expose custom geometry, semantic parts, articulation metadata, editable structure, and simulation-ready packaging within the same object-level representation.

We group representative prior systems by the asset strategy they adopt. The \emph{retrieved static asset} strategy is used by layout-oriented and retrieval-based scene synthesizers, including LayoutGPT~\citep{feng2023layoutgpt}, ATISS~\citep{paschalidou2021atiss}, InstructScene~\citep{lin2024instructscene}, Holodeck~\citep{yang2024holodeck}, LayoutVLM~\citep{sun2025layoutvlm}, and HSM~\citep{pun2025hsm}, which place pre-built static meshes drawn from curated 3D libraries. The \emph{retrieved articulated asset} and \emph{generated opaque mesh} strategies are both adopted by SceneSmith~\citep{pfaff2026scenesmith}, which retrieves articulated objects from PartNet-Mobility-style datasets~\citep{xiang2020sapien} for interactable elements and uses an image-to-3D pipeline to generate the remaining single-piece meshes. Table~\ref{tab:scenecode_vs_asset_strategies} compares these strategies along the axes most relevant to interactable scene synthesis.

\begin{table}[h]
    \centering
    \scriptsize
    \setlength{\tabcolsep}{2pt}
    \renewcommand{\arraystretch}{1.18}
    \begin{tabular}{p{0.24\linewidth} p{0.12\linewidth} p{0.17\linewidth} p{0.15\linewidth} p{0.09\linewidth} p{0.14\linewidth}}
    \toprule
    \textbf{Asset strategy} & \textbf{Custom geometry} & \textbf{Part hierarchy} & \textbf{New articulated objects} & \textbf{Editable} & \textbf{Simulation packaging} \\
    \midrule
    Retrieved static asset & Limited & Usually no & No & Limited & Sometimes \\
    Retrieved articulated asset & No & Yes, if dataset provides & No & Limited & Yes \\
    Generated opaque mesh & Yes & No / weak & No & Hard & Hard \\
    \textbf{\textsc{SceneCode} asset} & \textbf{Yes} & \textbf{Yes} & \textbf{Yes} & \textbf{Yes} & \textbf{Yes} \\
    \bottomrule
    \end{tabular}
    \caption{Comparison between \textsc{SceneCode} assets and common prior asset strategies. The key distinction is not the asset source, but the representation of the asset layer: \textsc{SceneCode} stores objects as executable, part-decomposed programs with explicit geometry and, when needed, articulation and simulation metadata.}
    \label{tab:scenecode_vs_asset_strategies}
\end{table}

The first two strategies are constrained by what already exists in an asset collection. Static retrieval can provide visually plausible objects, but usually lacks semantic part structure and cannot introduce new articulated object designs. Articulated retrieval preserves joints only when a suitable dataset asset is available, and its geometry is typically not customizable beyond coarse scaling or material edits. Opaque mesh generation can create new shapes, but the resulting asset is difficult to edit, inspect, decompose into functional parts, or compile into a simulator object with reliable links and joints.

By contrast, a \textsc{SceneCode} asset keeps geometry construction, part hierarchy, and packaging metadata coupled in the same executable object program. This allows the system to synthesize previously unavailable object variants, preserve editable substructure, mark movable components before export, and package the resulting object for downstream simulation rather than treating simulation support as a separate post-processing step.

\paragraph{Discussion: \textsc{SceneCode} versus SceneSmith.}
Among the agentic, simulation-ready scene generators, SceneSmith~\citep{pfaff2026scenesmith} is the closest in spirit to \textsc{SceneCode}: both target physically interactable indoor environments rather than purely visual layouts. The two systems, however, differ fundamentally in how they obtain object-level interaction.
SceneSmith composes a scene by mixing two asset sources: it \emph{retrieves} articulated objects from curated datasets (e.g.\ PartNet-Mobility-style libraries~\citep{xiang2020sapien}) and \emph{generates} the remaining objects as opaque meshes through an image-to-3D pipeline. Consequently, every interactable object in a SceneSmith scene is bounded by the coverage of the underlying articulated-object dataset; novel articulated categories or unusual joint configurations cannot be introduced on demand. The image-to-3D branch, in turn, produces single-piece meshes that lack semantic part decomposition, joint annotations, and code-level editability, so they cannot be promoted to interactable objects without a separate post-hoc articulation step.
\textsc{SceneCode} eliminates this split. Each object is synthesized as an executable program that simultaneously constructs geometry from primitives, decomposes the result into named semantic parts, attaches materials, and---when applicable---declares prismatic or revolute joints, collision proxies, and physical attributes inside the same program. This yields four concrete advantages over SceneSmith:
\begin{itemize}
    \item \textbf{Unbounded articulated coverage.} New interactable objects (e.g.\ a custom multi-drawer cabinet with a hinged top compartment) can be generated directly from the prompt, rather than being constrained to assets that already exist in an articulated-object dataset.
    \item \textbf{Intrinsic part hierarchy and articulation.} Parts and joints emerge as first-class elements of the program at generation time, instead of being either inherited from a curated dataset or absent (as in image-to-3D meshes).
    \item \textbf{Code-level editability and local regeneration.} Because every object is a program, individual attributes (dimensions, material, joint range, number of drawers) can be edited and re-executed locally, whereas SceneSmith's retrieved or image-to-3D assets are largely fixed once selected.
    \item \textbf{Unified simulation packaging.} Simulation metadata---transforms, support surfaces, bounding boxes, articulation files---is emitted by the same program and registered into a persistent scene state, removing the need to maintain separate retrieval, generation, and articulation pipelines.
\end{itemize}
In short, SceneSmith treats interactable assets as objects to be \emph{selected} (articulated retrieval) or \emph{shaped} (image-to-3D), while \textsc{SceneCode} treats them as objects to be \emph{programmed}, which is what enables on-demand, customizable, and editable interactable scene synthesis.

\section{More Analysis}
\label{app:more_analysis}

This appendix expands the per-metric discussion of the room-level results in Section~\ref{sec:exp_scene}.
We first detail the metrics where {SceneCode} is the leading method, then turn to the metrics where it is not and the design trade-offs that explain those gaps.

\paragraph{Semantic fidelity (CNT, ATR).}
{SceneCode} achieves the best CNT (0.7951) and the best ATR (0.7344), and is the only method that tops both semantic-fidelity metrics simultaneously.
The CNT margin over the strongest baseline SceneSmith is small (+0.0068), but the gap widens steadily over LayoutVLM (+0.1528) and HSM (+0.2623), showing that {SceneCode}'s per-object request pipeline reliably delivers the categories and counts named in the prompt.
The ATR advantage is more decisive: {SceneCode} surpasses SceneSmith by 0.0182, HSM by 0.1429, and LayoutVLM by a striking 0.4236.
We attribute this gain to the executable object program: per-object code receives attribute slots (color, material, style, dimensions) directly from the planner and realizes them at construction time, so descriptive modifiers in the prompt translate into actual asset properties rather than being approximated by the closest retrieved mesh.
This is the central reason {SceneCode} matches the prompt's atmosphere and material cues in Figure~\ref{fig:scene_results} more faithfully than the retrieval- or motif-based baselines.

\paragraph{Physical plausibility (NAV, COL, OOB).}
{SceneCode} also leads on all three implicit physical-plausibility metrics: NAV (0.9999), COL (0.0934), and OOB (0.0042).
The collision rate is roughly half of every baseline (HSM 0.1882, SceneSmith 0.1812, LayoutVLM 0.2105), and the out-of-bounds fraction is more than an order of magnitude lower than LayoutVLM (0.0592) and HSM (0.0722).
We attribute this to the executable object code emitting clean, axis-aligned bounding-box-faithful geometry that the placer can reason about precisely, so collision and floor-containment checks act on the same shape that is rendered.

\paragraph{Inter-object and architectural relations (OOR, OAR).}
SceneSmith leads on OOR (0.3723) and OAR (0.7241); {SceneCode} is a close second on OOR (0.3472, gap of 0.0251) and trails on OAR (0.6078).
A practical reason for this gap is the asset-acceptance behaviour of our pipeline: when an object program fails the ObjectPlan check beyond the retry budget, the agent skips that object rather than committing a low-quality asset, which removes some of the relation-bearing objects from the final scene.
Mesh-generation baselines such as SceneSmith do not face this trade-off because their image-to-3D branch reliably returns a visually plausible mesh for almost any request, even when fine-grained semantic accuracy is sacrificed.
We therefore view OOR as a near-tie and OAR as an actionable target for future work, e.g., adding a relation-aware reward to the placement loop and relaxing the asset-acceptance criterion for objects whose role is primarily relational.

\paragraph{Per-baseline trade-offs (SUP, ACC, OPC) and overall.}
The baseline-leading metrics each correspond to a focused design choice rather than overall scene quality.
HSM tops SUP (0.7072) because its hierarchical motifs hard-code on-support placement, but it ranks last on CNT and second-to-last on the relation metrics.
LayoutVLM tops ACC (0.9330) because its differentiable layout objective explicitly optimizes accessibility, and because the scenes it produces are notably sparse---the unnaturally empty rooms make most functional sides trivially unobstructed and effectively inflate the accessibility score---while collapsing on ATR (0.3108) and SUP (0.2566).
SceneSmith tops OPC (0.0792); LayoutVLM does not implement OPC at all (N/A).
{SceneCode} trails on SUP (the placer does not yet classify support type per object) and on OPC (no opening-clearance check is enforced); both are localized weaknesses tied to specific checks that can be added without changing the asset representation.

\section{Full User Study Results}
\label{app:user_study}

\begin{figure}[t]
  \centering
  \includegraphics[width=\linewidth]{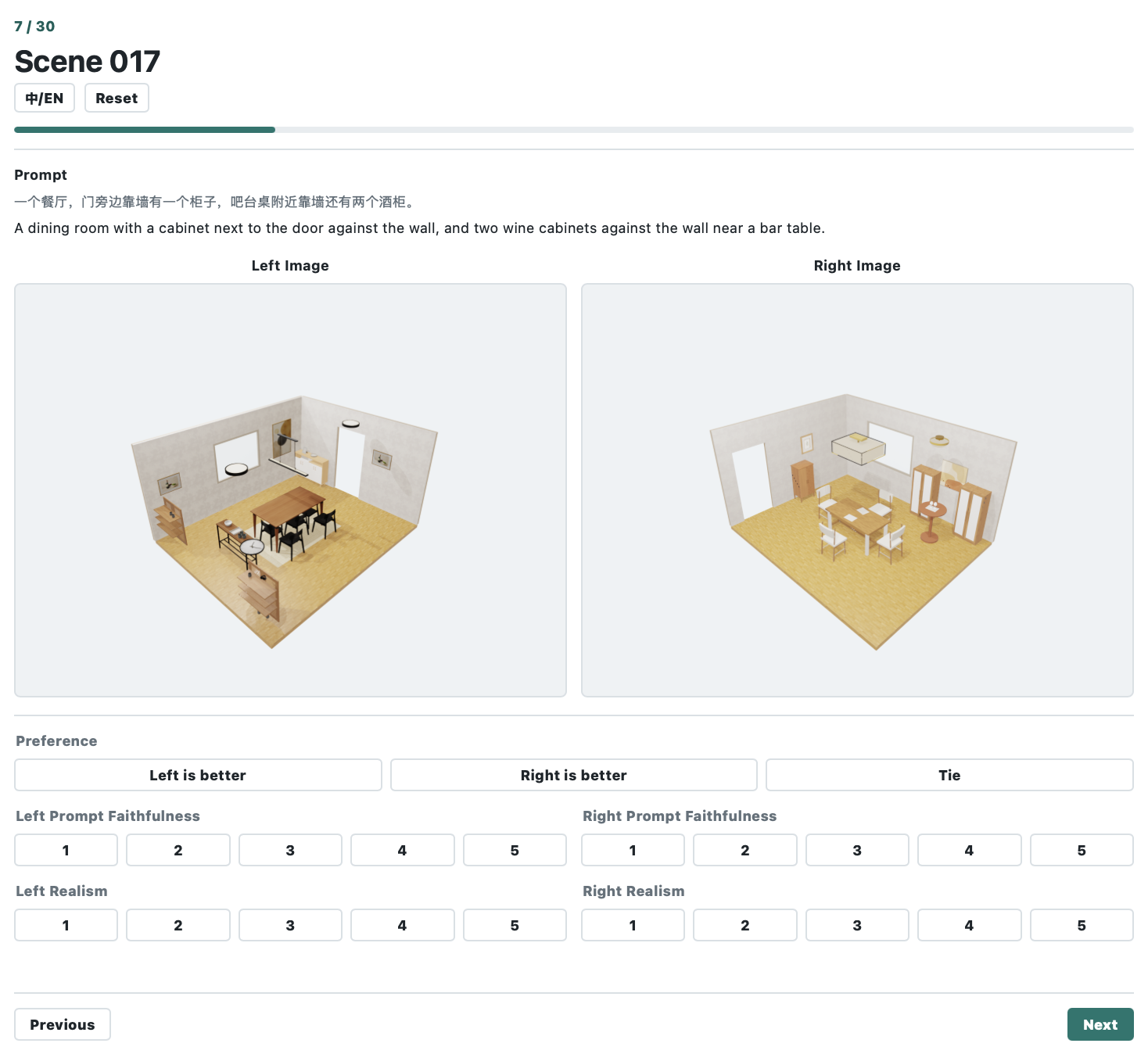}
  \caption{\textbf{User-study interface.} Screenshot of the UI used in our user study.}
  \label{fig:user_study_ui}
\end{figure}

\paragraph{Setup recap.}
The user study involves nine participants split evenly into three groups of three.
Group A compares {SceneCode} against SceneSmith, Group B against {HSM}, and Group C against LayoutVLM, on the same 30 SceneEval-100 room-level prompts used for the automatic evaluation.
For each prompt and each comparison, participants make a pairwise preference choice and rate both scenes on prompt faithfulness and realism using a 1--5 Likert scale.
Because {SceneCode} is rated by all nine participants while each baseline is rated only by the three participants in its own group, absolute means are not directly comparable across baselines.
We therefore report (i) the preference loss rate, defined as the fraction of comparisons in which the baseline is preferred over {SceneCode}, and (ii) the paired per-trial differences $\Delta\text{PF}$ and $\Delta\text{Realism}$: for each trial we compute $\Delta = \text{SceneCode} - \text{Baseline}$ on the same prompt and the same rater, then report the mean and a 95\% confidence interval over the $n{=}90$ paired trials per comparison.
This paired-difference design controls for rater identity and prompt sampling and isolates the comparison between {SceneCode} and each baseline.
Figure~\ref{fig:user_study_ui} shows the interface used for collecting the user-study ratings and pairwise preferences.

\begin{table}[t]
  \centering
  \caption{\textbf{Full user-study results.} Pairwise preference loss rate (fraction of trials where the baseline is preferred over {SceneCode}; lower is better) and paired per-trial differences $\Delta\text{PF}$ and $\Delta\text{Realism}$ (computed as $\text{SceneCode} - \text{Baseline}$ within each trial, then averaged), with 95\% confidence intervals over $n{=}90$ trials per comparison shown in smaller font.}
  \label{tab:user_study_full}
  \setlength{\tabcolsep}{6pt}
  \begin{tabular}{llcrr}
    \toprule
    Group & Baseline & Loss Rate $\downarrow$ & $\Delta$ PF $\uparrow$ & $\Delta$ Realism $\uparrow$ \\
    \midrule
    A & SceneSmith & 36.7\% & $+0.14$\,{\scriptsize $[-0.05,\,0.34]$} & $-0.34$\,{\scriptsize $[-0.56,\,-0.13]$} \\
    B & {HSM}      & 20.0\% & $+0.66$\,{\scriptsize $[\phantom{-}0.44,\,0.87]$} & $+0.27$\,{\scriptsize $[\phantom{-}0.06,\,0.47]$} \\
    C & LayoutVLM  &  3.3\% & $+1.23$\,{\scriptsize $[\phantom{-}0.94,\,1.52]$} & $+1.93$\,{\scriptsize $[\phantom{-}1.70,\,2.16]$} \\
    \bottomrule
  \end{tabular}
\end{table}

\paragraph{Pairwise preference.}
Table~\ref{tab:user_study_full} shows that the baseline-preferred fraction scales inversely with how competitive the baseline is on automatic metrics: only $3.3\%$ against LayoutVLM, $20.0\%$ against HSM, but $36.7\%$ against SceneSmith, the strongest retrieval-based baseline whose image-to-3D branch produces visually rich meshes for nearly any request.
This ordering is consistent with the $\Delta$PF column in Table~\ref{tab:room_level_results}: the smaller the prompt-faithfulness gap, the closer the preference becomes.

\paragraph{Realism trade-off.}
{SceneCode} wins on realism against HSM ($+0.27$) and LayoutVLM ($+1.93$), but loses to SceneSmith ($-0.34$).
This mirrors the OOR/OAR trade-off discussed above: SceneSmith's image-to-3D branch returns photo-realistic retrieved meshes for almost any request, whereas {SceneCode} prioritizes attribute-faithful programmatic geometry, which can read as less photoreal on visually rich items even when it is more prompt-faithful (note that {SceneCode} still wins on $\Delta$PF in the same group, $+0.14$, and on the automatic ATR metric).
The realism gap is therefore a property of the asset representation rather than of the planning pipeline, and is the natural target for a future neural texturing or material-refinement stage layered on top of the existing object programs.

\paragraph{Why per-group $\Delta$.}
A single global ranking would mix three different rater groups and three different baseline distributions, making any difference between baselines confounded with which participants happened to rate which method.
By holding the rater group fixed within each row of Table~\ref{tab:user_study_full} and only contrasting {SceneCode} against the baseline that group sees, the paired-trial $\Delta$ measures exactly what a participant in that group thought of {SceneCode} versus that one baseline on the same prompt.

\clearpage

\section{Computational Cost Statistics}
\label{app:cost_statistics}

We report the computational cost statistics for the same batch of evaluation runs used in our experiments.
Table~\ref{tab:cost_statistics_usd} summarizes token-based monetary cost in USD, and Table~\ref{tab:cost_statistics_time} summarizes wall-clock runtime.
The average object-construction and agent costs decompose the average token cost.

\begin{table}[h]
    \centering
    \small
    \renewcommand{\arraystretch}{1.18}
    \begin{tabular}{lcc}
    \toprule
    \textbf{Cost item} & \textbf{USD} & \textbf{Ratio} \\
    \midrule
    Total token cost & \$630.188234 & N/A \\
    Average token cost & \$21.730629 & 100.0000\% \\
    Average object-construction cost & \$13.866764 & 61.6417\% \\
    Average agent cost & \$7.863865 & 38.3583\% \\
    \bottomrule
    \end{tabular}
    \caption{Token-based monetary cost statistics for the evaluation runs. The average object-construction cost and average agent cost decompose the average token cost.}
    \label{tab:cost_statistics_usd}
\end{table}

\begin{table}[h]
    \centering
    \small
    \renewcommand{\arraystretch}{1.18}
    \begin{tabular}{lcc}
    \toprule
    \textbf{Statistic} & \textbf{Wall-clock time} & \textbf{Seconds} \\
    \midrule
    Average time cost & 7:25:36 & 26736.5 s \\
    Maximum time cost & 16:39:44 & 59984.7 s \\
    Minimum time cost & 2:06:18 & 7578.1 s \\
    \bottomrule
    \end{tabular}
    \caption{Wall-clock time cost statistics for the evaluation runs, reported both in hours:minutes:seconds and seconds.}
    \label{tab:cost_statistics_time}
\end{table}

\clearpage

\section{Additional Qualitative Demonstrations}
\label{app:more_demos}

\subsection{Scene Demo}
\label{app:more_scene_demos}
\begin{center}
  \centering
  \includegraphics[width=\linewidth,height=0.80\textheight,keepaspectratio]{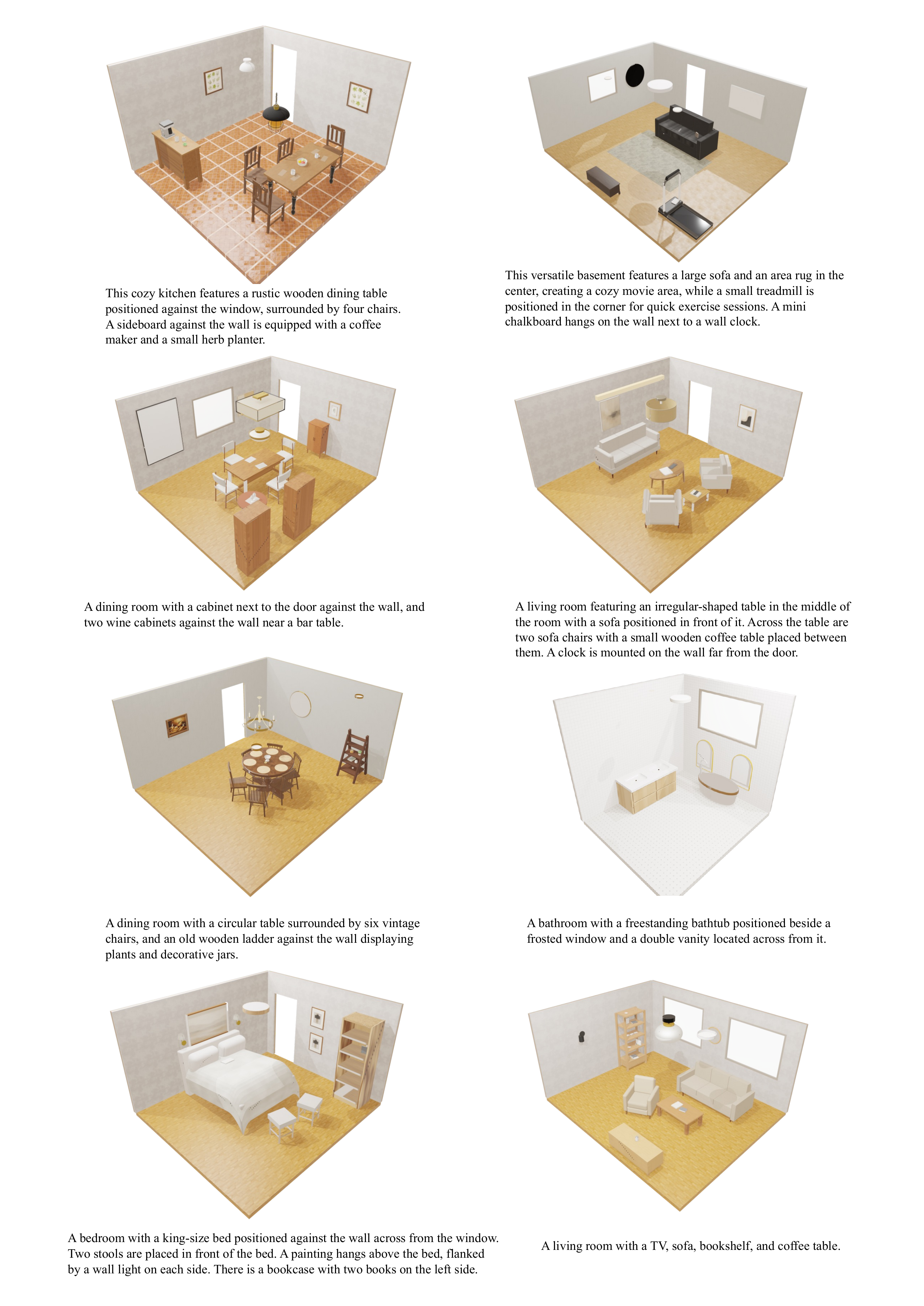}
  \captionof{figure}{\textbf{Additional room-level demonstrations.}
  \textsc{SceneCode} generates kitchen, basement, dining room, living room, bathroom, and bedroom scenes with prompt-faithful object coverage, spatial layout, and stylistic attributes.}
  \label{fig:more_scene_demos}
\end{center}

Figure~\ref{fig:more_scene_demos} provides additional qualitative evidence for the room-level behavior of \textsc{SceneCode}.
Across these prompts, the generated rooms contain the requested furniture and functional objects, place them in plausible room-scale configurations, and retain attribute cues such as rustic dining furniture, exercise equipment in a basement corner, bathroom fixtures near windows, and decorative wall objects.
These examples complement the main-paper comparison by showing that the code-driven representation is not limited to isolated showcase scenes: the planner can instantiate many object requests, arrange them coherently, and preserve enough free space for navigation and physical use.

\subsection{Object Demo}

\begin{center}
  \centering
  \includegraphics[width=\linewidth]{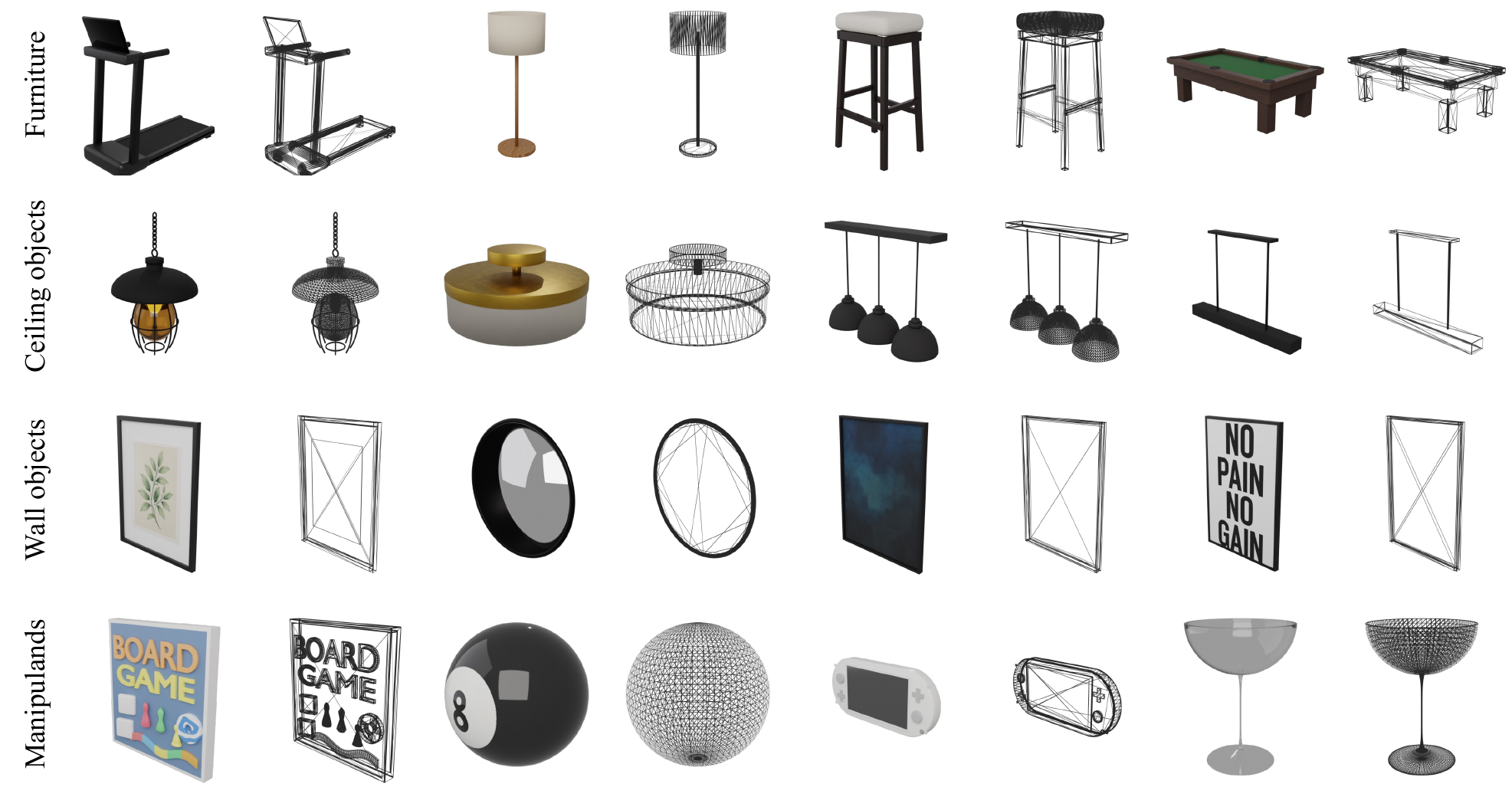}
  \captionof{figure}{\textbf{Additional object-level demonstrations.}
  Rendered objects and corresponding wireframes for furniture, ceiling objects, wall objects, and manipulands generated by \textsc{SceneCode}.
  The examples show that code-generated assets can express materials, numbers, text, and nontrivial geometric structure while preserving explicit mesh organization.}
  \label{fig:object_demo_gallery}
\end{center}

Figure~\ref{fig:object_demo_gallery} highlights the object-level advantage of generating assets as executable construction programs.
The rendered views demonstrate visually distinct materials and appearances, while the paired wireframes expose clean, editable geometric structure rather than an opaque reconstructed mesh.
This structure is useful for downstream editing and simulation because individual components, dimensions, and material assignments remain accessible in code.
The wall-art examples show that \textsc{SceneCode} can place image-like and text-like content on thin framed objects; the billiard ball illustrates number and material control on curved glossy geometry; and the board-game box shows that text, colored graphics, and layered physical structure can be generated together.
Together, these cases support the central claim that \textsc{SceneCode} improves scene synthesis not only by placing objects, but by producing controllable code-generated assets that remain interpretable, editable, and compatible with physical scene construction.

\subsection{Articulated Object Demo}

\begin{center}
  \centering
  \includegraphics[width=\linewidth]{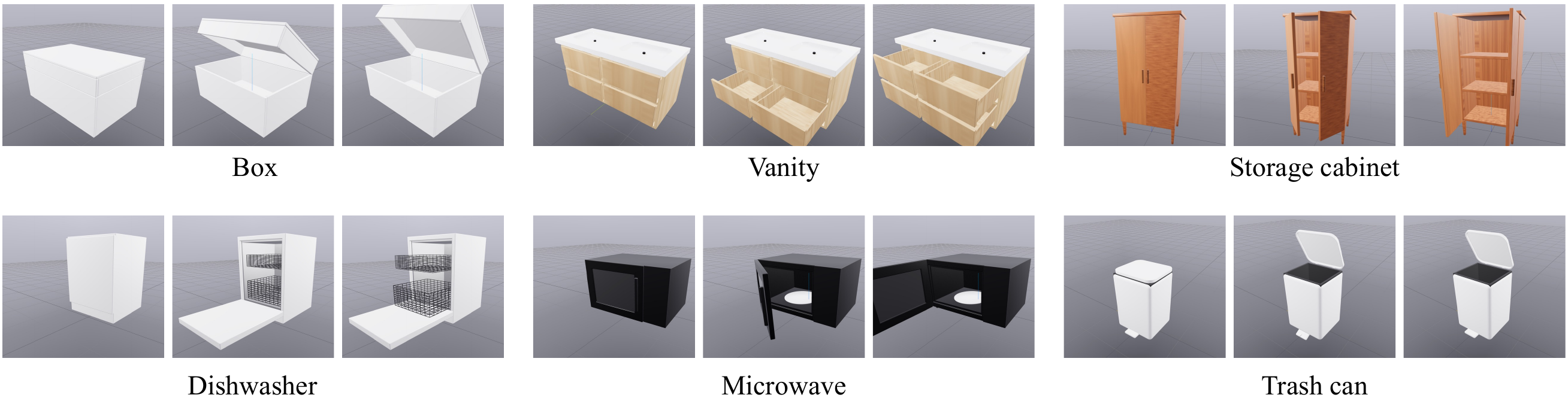}
  \captionof{figure}{\textbf{Articulated object demonstrations.}
  \textsc{SceneCode} generates articulated household objects with correct geometric structure, storage interiors such as shelves and partitions, and movable components that remain available for downstream interaction.}
  \label{fig:articulation_object_demo}
\end{center}

Figure~\ref{fig:articulation_object_demo} demonstrates that \textsc{SceneCode} can synthesize articulated household objects as structured executable assets.
The generated storage objects are not only plausible from the exterior, but also preserve meaningful internal geometry such as compartments, shelves, and dividers.
Their articulation is represented through independent functional parts, including doors, drawers, and lids, so the resulting assets can support physically meaningful opening or sliding behavior rather than remaining static meshes.
Because these objects are generated from code conditioned on the request, their dimensions, internal layout, materials, and joint configuration can be customized on demand without retrieving a nearest neighbor from a fixed asset library or relying on laborious manual modeling.

\section{Executable Object-Code Demonstration}
\label{app:code}

\lstdefinestyle{scenecode}{
    basicstyle=\ttfamily\scriptsize,
    numbers=left,
    numberstyle=\tiny\color{gray},
    stepnumber=1,
    numbersep=6pt,
    frame=single,
    rulecolor=\color{black!18},
    backgroundcolor=\color{black!2},
    breaklines=true,
    breakatwhitespace=false,
    columns=fullflexible,
    keepspaces=true,
    showstringspaces=false,
    tabsize=2,
    captionpos=b,
    xleftmargin=1.0em,
    framexleftmargin=1.0em,
    keywordstyle=\color{blue!60!black},
    commentstyle=\color{green!35!black},
    stringstyle=\color{red!55!black}
}

\noindent
\begin{minipage}[c]{0.2\linewidth}
    \centering
    \includegraphics[width=\linewidth]{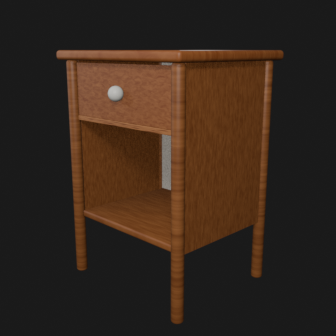}
    \captionof{figure}{Generated nightstand demo.}
    \label{fig:nightstand_code_demo}
\end{minipage}\hfill
\begin{minipage}[c]{0.75\linewidth}
    This appendix provides a concrete example of the executable object representation used by
    \textsc{SceneCode}. We use a generated nightstand as the running example. The object-level
    Blender program constructs the movable drawer from geometric primitives and procedural
    materials, while the exported SDF preserves the drawer as an independent simulation link
    with a prismatic joint.
\end{minipage}
\vspace{0.6em}

\subsection{Blender Drawer Construction Program}
\label{app:blender_drawer_demo}

Listing~\ref{lst:drawer_blender_code} shows the complete Blender Python program for the
drawer part. The code exposes the semantic substructure of the drawer---front panel, side
walls, back wall, bottom panel, and knob---rather than storing the object as an opaque mesh.
It also records construction dimensions, positions, materials, beveling, and the local origin
used by downstream packaging.

\begin{lstlisting}[
    style=scenecode,
    language=Python,
    caption={Complete Blender Python demo for the generated drawer part.},
    label={lst:drawer_blender_code}
]
# Part: drawer
# Description: A pull-out drawer located in the upper section of the cabinet.
# Generated by Part Constructor Agent

import bpy
import math

def srgb_to_linear(srgb_color, alpha=1.0):
    linear_color = []
    for c in srgb_color:
        if c <= 0.04045:
            linear_color.append(c / 12.92)
        else:
            linear_color.append(((c + 0.055) / 1.055) ** 2.4)
    return tuple(linear_color) + (alpha,)

def create_drawer():
    """
    Create part: drawer
    
    Shape: cube
    Dimensions: 0.56 x 0.46 x 0.22
    Position: (0.0, -0.03, 0.84)
    """
    
    part_name = "drawer"
    
    # Remove same-name object if it already exists
    if part_name in bpy.data.objects:
        bpy.data.objects.remove(bpy.data.objects[part_name], do_unlink=True)
        
    # === Materials Setup ===
    # 1. Wood Material for drawer box and front panel
    wood_mat_name = "drawer_wood_material"
    wood_mat = bpy.data.materials.get(wood_mat_name)
    if wood_mat is None:
        wood_mat = bpy.data.materials.new(name=wood_mat_name)
        wood_mat.use_nodes = True
        nodes = wood_mat.node_tree.nodes
        links = wood_mat.node_tree.links
        
        bsdf = next((n for n in nodes if n.type == 'BSDF_PRINCIPLED'), None)
        if bsdf:
            bsdf.inputs["Roughness"].default_value = 0.4
            bsdf.inputs["Metallic"].default_value = 0.0
            
            # Wood grain procedural texture
            tex_noise = nodes.new(type='ShaderNodeTexNoise')
            tex_noise.inputs["Scale"].default_value = 20.0
            tex_noise.inputs["Detail"].default_value = 4.0
            tex_noise.inputs["Distortion"].default_value = 1.0
            
            mix = nodes.new(type='ShaderNodeMix')
            mix.data_type = 'RGBA'
            mix.blend_type = 'MULTIPLY'
            mix.inputs["Factor"].default_value = 0.2
            
            base_color_srgb = (0.75, 0.5, 0.3)
            darker_wood_srgb = (0.6, 0.35, 0.2)
            
            mix.inputs[6].default_value = srgb_to_linear(base_color_srgb) # A: Base Color
            mix.inputs[7].default_value = srgb_to_linear(darker_wood_srgb) # B: Texture Color
            
            links.new(tex_noise.outputs["Fac"], mix.inputs["Factor"])
            links.new(mix.outputs[2], bsdf.inputs["Base Color"])

    # 2. Plastic Material for drawer knob
    knob_mat_name = "drawer_knob_material"
    knob_mat = bpy.data.materials.get(knob_mat_name)
    if knob_mat is None:
        knob_mat = bpy.data.materials.new(name=knob_mat_name)
        knob_mat.use_nodes = True
        nodes = knob_mat.node_tree.nodes
        
        bsdf = next((n for n in nodes if n.type == 'BSDF_PRINCIPLED'), None)
        if bsdf:
            bsdf.inputs["Base Color"].default_value = srgb_to_linear((0.95, 0.95, 0.92))
            bsdf.inputs["Roughness"].default_value = 0.3
            bsdf.inputs["Metallic"].default_value = 0.0

    parts_to_join = []

    # Helper function to create box-like sub-parts
    def add_sub_part_cube(name, dimensions, position, material):
        bpy.ops.mesh.primitive_cube_add(size=1, location=position)
        obj = bpy.context.active_object
        obj.name = name
        obj.scale = dimensions
        bpy.ops.object.transform_apply(scale=True)
        if material:
            obj.data.materials.append(material)
        parts_to_join.append(obj)
        return obj

    # === Create Sub-parts ===
    
    # 1. Drawer Front Panel
    add_sub_part_cube(
        name="drawer_front_panel",
        dimensions=(0.56, 0.02, 0.22),
        position=(0.0, -0.25, 0.84),
        material=wood_mat
    )
    
    # 2. Drawer Left Wall
    add_sub_part_cube(
        name="drawer_left_wall",
        dimensions=(0.015, 0.44, 0.18),
        position=(-0.2725, -0.02, 0.84),
        material=wood_mat
    )
    
    # 3. Drawer Right Wall
    add_sub_part_cube(
        name="drawer_right_wall",
        dimensions=(0.015, 0.44, 0.18),
        position=(0.2725, -0.02, 0.84),
        material=wood_mat
    )
    
    # 4. Drawer Back Wall
    add_sub_part_cube(
        name="drawer_back_wall",
        dimensions=(0.53, 0.015, 0.18),
        position=(0.0, 0.1925, 0.84),
        material=wood_mat
    )
    
    # 5. Drawer Bottom
    add_sub_part_cube(
        name="drawer_bottom",
        dimensions=(0.53, 0.44, 0.015),
        position=(0.0, -0.02, 0.7575),
        material=wood_mat
    )
    
    # 6. Drawer Knob (Sphere)
    bpy.ops.mesh.primitive_uv_sphere_add(radius=0.03, location=(0.0, -0.28, 0.84))
    knob = bpy.context.active_object
    knob.name = "drawer_knob"
    knob.data.materials.append(knob_mat)
    parts_to_join.append(knob)

    # === Merge all sub-parts into a single object ===
    bpy.ops.object.select_all(action='DESELECT')
    for p in parts_to_join:
        p.select_set(True)
        
    bpy.context.view_layer.objects.active = parts_to_join[0]
    bpy.ops.object.join()
    
    obj = bpy.context.active_object
    obj.name = part_name
    obj.data.name = f"{part_name}_mesh"
    
    # === Apply modifiers ===
    bevel = obj.modifiers.new(name="Bevel", type='BEVEL')
    bevel.width = 0.002
    bevel.segments = 3
    # Limit bevel to a certain angle to avoid messing up the spherical knob geometry
    bevel.limit_method = 'ANGLE'
    bevel.angle_limit = math.radians(30)
    
    # === Set Origin to the Drawer's Main Position ===
    # This ensures rotation and movement pivot from the correct center
    saved_cursor = bpy.context.scene.cursor.location.copy()
    bpy.context.scene.cursor.location = (0.0, -0.03, 0.84)
    bpy.ops.object.origin_set(type='ORIGIN_CURSOR')
    bpy.context.scene.cursor.location = saved_cursor

    return obj

# Execute creation
if __name__ == "__main__":
    obj = create_drawer()
    print(f"Created {obj.name}: {obj.name}")
\end{lstlisting}

\subsection{SDF Drawer Link and Articulation}
\label{app:sdf_drawer_demo}

The corresponding SDF export keeps the drawer as its own link. Listing~\ref{lst:sdf_drawer_link}
shows the drawer link header, inertial properties, visual mesh, and the first collision proxy
directly from the generated SDF file. The remaining drawer collision entries follow the same
schema in the source file and are omitted here for compactness.

\begin{lstlisting}[
    style=scenecode,
    language=XML,
    firstline=332,
    lastline=371,
    firstnumber=332,
    caption={Drawer link excerpt from \texttt{demo/nightstand\_1777368137052.sdf}.},
    label={lst:sdf_drawer_link}
]
    <link name="drawer">
      <pose relative_to="base_link">0.00000000 -0.02083333 0.58333333 0.00000000 0.00000000 0.00000000</pose>
      <inertial>
        <mass>3.500000</mass>
        <pose>0.000000 -0.013572 -0.020292 0 0 0</pose>
        <inertia>
          <ixx>5.539828e-02</ixx>
          <iyy>6.690811e-02</iyy>
          <izz>1.101321e-01</izz>
          <ixy>2.495866e-11</ixy>
          <ixz>-3.135577e-12</ixz>
          <iyz>1.457055e-03</iyz>
        </inertia>
      </inertial>
      <visual name="drawer_visual">
        <pose>0 0 0 0 0 0</pose>
        <geometry>
          <mesh>
            <uri>visual/drawer_visual.gltf</uri>
            <scale>0.6944444444444444 0.6944444444444444 0.6944444444444444</scale>
          </mesh>
        </geometry>
      </visual>
      <collision name="drawer_collision_0">
        <pose>0 0 0 0 0 0</pose>
        <geometry>
          <mesh>
            <uri>collision/drawer_collision_0.obj</uri>
            <scale>0.6944444444444444 0.6944444444444444 0.6944444444444444</scale>
          </mesh>
        </geometry>
        <surface>
          <friction>
            <ode>
              <mu>0.500</mu>
              <mu2>0.500</mu2>
            </ode>
          </friction>
        </surface>
      </collision>
\end{lstlisting}

Listing~\ref{lst:sdf_drawer_joint} shows the articulation metadata for the same drawer. The
prismatic joint connects the drawer link to the nightstand base, slides along the negative
\texttt{y} direction, and constrains the travel range to \texttt{[0.0, 0.4]}.

\begin{lstlisting}[
    style=scenecode,
    language=XML,
    firstline=601,
    lastline=617,
    firstnumber=601,
    caption={Prismatic drawer joint excerpt from \texttt{demo/nightstand\_1777368137052.sdf}.},
    label={lst:sdf_drawer_joint}
]
    <joint name="joint_drawer" type="prismatic">
      <parent>base_link</parent>
      <child>drawer</child>
      <axis>
        <xyz>0 -1 0</xyz>
        <limit>
          <lower>0.0</lower>
          <upper>0.4</upper>
          <effort>100</effort>
          <velocity>1.0</velocity>
        </limit>
        <dynamics>
          <damping>0.05</damping>
          <friction>0.05</friction>
        </dynamics>
      </axis>
    </joint>
\end{lstlisting}

\section{Limitations}
\label{app:limitations}

SceneCode focuses on physically interactable indoor scene synthesis, where objects are typically organized by strong functional and architectural priors such as floors, walls, support surfaces, furniture layouts, and household articulation patterns. While this setting already covers diverse household environments and a broad range of common objects, extending the same purely code-driven paradigm to outdoor or large-scale mixed environments remains an open direction. Outdoor scenes introduce larger spatial scales, irregular terrain, organic geometry, vegetation, complex illumination, weather effects, and less structured object layouts, which may require additional procedural priors and verification strategies beyond those used in our current indoor pipeline.

A second direction concerns the visual richness of code-generated assets. SceneCode prioritizes explicit part structure, editability, clean geometry, and simulation-oriented asset construction. This design makes object programs more usable and compatible with downstream physical interaction, but purely primitive-based programs may not always capture the fine-grained surface detail, or photorealistic texture variation provided by retrieval or image-to-3D pipelines. Future work could combine executable object programs with neural texture synthesis, material refinement, or detail-preserving geometry augmentation while keeping the underlying semantic parts and articulation metadata editable.

Finally, our implementation emphasizes generation quality, traceability, and execution-guided validation. This design is useful for producing editable world programs, but it also introduces nontrivial runtime due to program synthesis, execution, repair, and refinement. We expect future systems to improve throughput through parallel object generation, and specialized code-generation models distilled for 3D asset construction.



\end{document}